\documentclass[fleqn,10pt]{my_preprint}

\usepackage{subcaption}
\graphicspath{{.}} \DeclareGraphicsExtensions{.pdf, .png}

\usepackage{bm}
\usepackage{mathtools}
\usepackage{siunitx}
\usepackage{blindtext}
\usepackage{algorithm} 
\usepackage{algpseudocodex}
\usepackage[dvipsnames,table]{xcolor}

\newcommand{\boruvka}{Bor\r{u}vka}

\newcommand{\graph}[1]{\bm{#1}}
\renewcommand{\vec}[1]{\bm{\mathbf{#1}}}

\newcommand{\e}{\varepsilon}
\newcommand{\N}{\rm N}
\newcommand{\D}{\rm D}

\newcommand\SetSymbol[1][]{\nonscript\:#1\vert\allowbreak\nonscript\:\mathopen{}}
\providecommand\given{}
\DeclarePairedDelimiterX\Set[1]{\{}{\}}{\renewcommand\given{\SetSymbol[\delimsize]}#1}

\undef{\min} \undef{\max}
\DeclareMathOperator*{\setmin}{min}
\DeclareMathOperator*{\setmax}{max}
\DeclareMathOperator*{\setsdist}{d_s}
\DeclarePairedDelimiterXPP\min[1]{\operatorname{min}}{(}{)}{}{#1}
\DeclarePairedDelimiterXPP\max[1]{\operatorname{max}}{(}{)}{}{#1}
\DeclarePairedDelimiterXPP\argmin[1]{\operatorname{arg\,min}}{(}{)}{}{#1}
\DeclarePairedDelimiterXPP\argmax[1]{\operatorname{arg\,max}}{(}{)}{}{#1}
\DeclarePairedDelimiterXPP\dist[1]{\operatorname{d}}{(}{)}{}{#1}
\DeclarePairedDelimiterXPP\core[1]{\operatorname{\kappa}}{(}{)}{}{#1}
\DeclarePairedDelimiterXPP\mutreach[1]{\operatorname{d_{mut}}}{(}{)}{}{#1}
\DeclarePairedDelimiterXPP\sdist[1]{\operatorname{d_s}}{(}{)}{}{#1}
\DeclarePairedDelimiterXPP\ari[1]{\operatorname{ARI}}{(}{)}{}{#1}

\title{Persistent Multiscale Density-based Clustering}

\author[1]{Daniël Bot}
\author[2]{Leland McInnes}
\author[3]{Jan Aerts}
\affil[1]{UHasselt, Data Science Institute (DSI)}
\affil[2]{Tutte Institute for Mathematics and Computing}
\affil[3]{KU Leuven, Augmented Intelligence for Data Analytics Lab, Department of Biosystems}

\begin{abstract}
  Clustering is a cornerstone of modern data analysis. Detecting clusters in exploratory data analyses (EDA) requires algorithms that make few assumptions about the data. Density-based clustering algorithms are particularly well-suited for EDA because they describe high-density regions, assuming only that a density exists. Applying density-based clustering algorithms in practice, however, requires selecting appropriate hyperparameters, which is difficult without prior knowledge of the data distribution. For example, DBSCAN requires selecting a density threshold, and HDBSCAN* relies on a minimum cluster size parameter. In this work, we propose Persistent Leaves Spatial Clustering for Applications with Noise (PLSCAN), a multiscale density-based clustering algorithm that replaces HDBSCAN*'s fixed minimum cluster size pruning of a mutual-reachability linkage hierarchy with a persistence-based cluster selection procedure. Effectively, PLSCAN identifies all minimum cluster sizes for which HDBSCAN* produces stable (leaf) clusters. In concept, PLSCAN applies scale-space clustering principles and is equivalent to persistent homology on a novel metric space. We compare its performance to HDBSCAN* on several real-world datasets, demonstrating that it achieves a higher median ARI, is less sensitive to changes in the number of mutual reachability neighbours, and has higher stability under resampling. Additionally, we compare PLSCAN's computational costs to $k$-Means++, demonstrating competitive run-times on low-dimensional datasets. At higher dimensions, run times scale more similarly to HDBSCAN*.
\end{abstract}

\copyright{This work is licensed under a Creative Commons Attribution-NonCommercial-NoDerivatives 4.0 International License. Copyright may be transferred without notice, after which this version may no longer be accessible.}

\begin{document}
\maketitle{}%

\section{Introduction}%
\label{sec:introduction}%

Clustering is a cornerstone of modern data analysis~\citep{jain2010clustering}, with applications ranging from biology~\citep{peng2022cdc} and astronomy~\citep{tian2025progress} to marketing~\citep{alvesgomes2023marketing}. By revealing groups of similar observations, clustering algorithms serve as essential tools for a range of applications, including pattern discovery~\citep{rieck2016clustering} and data compression~\citep{krcal2015incremental} in increasingly complex datasets.

In this work, we focus on exploratory data analysis (EDA) as the primary use case. EDA aims to form effective descriptions of unfamiliar datasets to support understanding and reasoning~\citep{tukey1977exploratory}. This use case imposes several requirements on clustering methods. For example, algorithms should not rely on prior knowledge, such as the number, size, shape, or density of clusters. Similarly, clustering methods should have a minimal number of input parameters, thereby preventing practitioners from imposing their expectations on the results~\citep{keogh2004parameter-light}. The ability to reveal clusters at multiple levels of detail and be efficiently computable on large datasets is beneficial in practice.

\subsection{Related work}
\label{sec:introduction:related work}

While many clustering algorithms have been developed for a broad range of applications~\citep{ezugwu2022survey}, few algorithms fully address the needs of EDA. Clustering algorithms are typically categorised as hierarchical or partitional~\citep{jain2010clustering}, where the latter category is sometimes also referred to as model-based~\citep{fred2005lifetimes}. Partitional clustering algorithms include mixture decomposition methods~\citep{baraldi2002art}, prototype-based methods~\citep{liu2009prototype}, shape fitting techniques~\citep{wang2016fuzzy}, and $k$-Means~\citep{macqueen1967some,lloyd1982kmeans} and its variants~\citep{arthur2006kmeans,park2009medoids,bezdek1984fcm}. Hierarchical clustering algorithms can be further divided into agglomerative~\citep{elsonbaty1998agglomerative} and divisive~\citep{chavent1998monothetic} approaches. They rely on linkage methods to define similarity between data points and clusters~\citep[e.g.,][]{ward1963hierarchical}. Single and complete linkage are common, emphasising connectedness and compactness, respectively~\citep{fred2005lifetimes}.

\paragraph*{Other clustering paradigms}
Not all clustering algorithms fit in these categories. Spectral clustering projects data into a space that reflects its intrinsic structure~\citep{belkin2003laplacianeigenmaps} before extracting clusters~\citep{damle2019simple}. Boundary-seeking algorithms quantify the spread of nearest neighbours to separate weakly connected clusters~\citep{peng2022cdc}. Evolutionary and genetic algorithms employ stochastic optimisation to search for cost-minimising clusterings~\citep{xu2005survey, ezugwu2022survey}. Streaming clustering algorithms efficiently process new data, adapting to changes in the data distribution~\citep{sun2024tws}. Correlation constrained clustering algorithms identify clusters with strong linear~\citep{bohm20044c} or non-linear correlations~\citep{tung2005curler}.

\paragraph*{Density-based clustering}
\begin{figure*}[t]
  \centering
  \begin{subfigure}{0.33\textwidth}
    \centering
    \includegraphics[width=\textwidth]{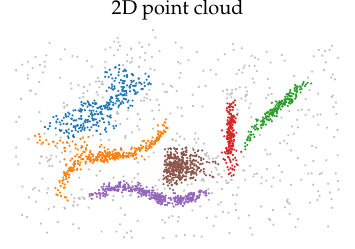}
    \caption{}%
    \label{fig:algorithm:smoothing:data}
  \end{subfigure}%
  \begin{subfigure}{0.66\textwidth}
    \centering
    \includegraphics[width=\textwidth]{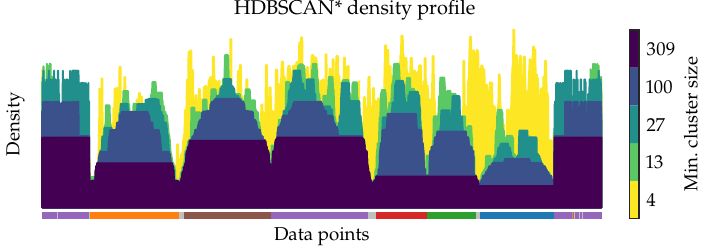}
    \caption{}%
    \label{fig:algorithm:smoothing:profile}
  \end{subfigure}%
  \hfill{}%
  \caption{OPTICS-like visualisation~\citep{ankerst1999optics} demonstrating the minimum cluster size parameter's smoothing effect. (\textbf{a}) A 2D point cloud from \citet{mcinnes2022documentation} with HDBSCAN* leaf-clusters for $k=5$ and $m_{\rm c} = 100$. (\textbf{b}) HDBSCAN*'s modelled density profile at multiple minimum cluster sizes. The $x$-axis contains all data points ordered to mimic the shape of probability density functions. The ordering is fixed for all minimum cluster size thresholds and does not encode proximity in data space. Colour encodes the minimum cluster size threshold. Higher thresholds smooth the modelled density profile by pruning small peaks. The color-strip at the bottom indicates points' cluster label.}%
  \label{fig:algorithm:smoothing}
\end{figure*}
Density-based clustering algorithms are particularly well-suited for EDA. They non-parametrically estimate the data's sampling distribution and detect clusters as modes in the resulting density profile~\citep{hartigan1975clustering}. Consequently, algorithms based on these principles describe the number, shape, and size of high-density regions present in the data. For example, DBSCAN~\citep{ester1996dbscan} identifies clusters as regions that exceed a density threshold $\lambda$ and contain at least $k$ points. Points with densities below $\lambda$ are rejected as noise. OPTICS~\citep{ankerst1999optics} visualises the estimated density profiles, supporting analysts in selecting $\lambda$ thresholds. HDBSCAN*~\citep{campello2013density,campello2015hdbscan} creates a cluster hierarchy over all densities and provides two cluster selection strategies to extract flat clusterings. 

\paragraph*{Hyperparameter selection}
Despite these advantages, extracting useful clusterings with density-based algorithms remains a challenging task because their parameters need to be tuned per application~\citep{keogh2004parameter-light}. For example, DBSCAN requires appropriate density thresholds. In practice, selecting these thresholds is difficult without prior knowledge of the data distribution~\citep{schubert2017dbscan}. Additionally, HDBSCAN* relies on a minimum cluster size parameter ($m_{\rm c}$). \citet{neto2021multiscale} facilitate the selection of the $m_{\rm c}$ parameter by explicitly computing hierarchies for $m_{\rm c} \in [1, m_{\rm c,max}]$. In their formulation, however, the number of mutual reachability neighbours $k$ is coupled to the pruning threshold $m_{\rm c}$ through $k = m_{\rm c}$. Consequently, both the mutual reachability distances and the resulting cluster hierarchy change across the explored scales, rather than only the minimum cluster size pruning. They report their algorithm can compute more than one hundred hierarchies in the time it takes to compute HDBSCAN* twice~\citep{neto2021multiscale}.

\paragraph*{Scale-space clustering}
HDBSCAN* uses its minimum cluster size parameter to prune a linkage hierarchy, thereby smoothing the modelled density profile by removing smaller peaks (Fig.~\ref{fig:algorithm:smoothing}). Scale-space clustering also applies smoothing operations to find clusters that match human perception~\citep{leung2000scalespace}. For example, \citet{leung2000scalespace} apply Gaussian filters to probability density functions and track the resulting modes as the kernel bandwidth increases to create a cluster hierarchy. They argue that clusters' (logarithmic) bandwidth-lifetimes are a measure of their importance, based on the idea that clusters should be perceivable over a wide range of scales~\citep{leung2000scalespace}. 

The work by \citet{leung2000scalespace} is designed around 2D continuous functions, and their demonstrations were limited to datasets represented as images. Additional studies have applied scale-space clustering to other spaces~\citep{hirano2015scalespace,wang2017salient} and discussed its theoretical limitations~\citep{florack2000topological,loog2001scalespace}.

\subsection{Contributions}
\label{sec:introduction:contributions}

In the present paper, we propose \emph{Persistent Leaves Spatial Clustering for Applications with Noise} (PLSCAN), a density-based clustering algorithm designed for exploratory data analysis. PLSCAN combines HDBSCAN*'s mutual-reachability linkage hierarchy construction with a novel persistence-based multiscale minimum cluster size cluster selection strategy. We apply scale-space clustering by varying HDBSCAN*'s minimum cluster size parameter, effectively smoothing the modelled density profile. PLSCAN improves on \citet{neto2021multiscale} by efficiently finding all minimum cluster sizes at which HDBSCAN*'s leaf clusters change. In other words, where HDBSCAN* provides cluster hierarchies describing DBSCAN clusters over all density scales, our algorithm provides cluster hierarchies containing HDBSCAN* leaf-clusters over all minimum cluster sizes. We demonstrate that this operation is equivalent to computing zero-dimensional persistent homology using a novel distance metric, thereby highlighting the approach's generalisability to other linkage tree pruning methods.

In summary, we make the following contributions:
\begin{itemize}
 \item A parameter-light density-based clustering algorithm suitable for exploratory data analysis.
 \item An efficient algorithm for computing minimum cluster size filtrations and a \emph{leaf tree} data structure modelling the resulting hierarchy of local density maxima over all minimum cluster sizes.
 \item A novel cluster selection strategy identifying \emph{cluster layers} with persistent clusters at multiple levels of detail.
 \item An open-source implementation available at: \url{https://github.com/jelmerbot/fast_plscan}.
\end{itemize}
Like HDBSCAN*, PLSCAN selects clusters at varying densities, rejects noise points in low-density regions, and produces cluster hierarchies that reflect all potential clusterings in a dataset.

\subsection{Organisation}%
\label{sec:introduction:organisation}

The remainder of the present paper is organised as follows: Section~\ref{sec:background:hdbscan} provides background information on density-based clustering with HDBSCAN*. Section~\ref{sec:maths} relates our approach to mathematical concepts from algebraic topology. Section~\ref{sec:algorithm} presents the PLSCAN algorithm from a computational perspective. Section~\ref{sec:demo} evaluates PLSCAN on real-world datasets, comparing its parameter sensitivity, stability under resampling, and computational costs with those of HDBSCAN* and $k$-Means. Finally, Section~\ref{sec:discussion} concludes the paper and outlines directions for future work.

\section{Background: HDBSCAN*}%
\label{sec:background:hdbscan}

Density-based clustering treats modes in probability density functions as clusters~\citep{hartigan1975clustering}. HDBSCAN* applies these principles to high-dimensional point clouds, estimating their density profile non-parametrically and constructing a corresponding cluster hierarchy~\citep{campello2015hdbscan}. This section provides a brief overview of HDBSCAN*. We refer to \citet{campello2015hdbscan} and \citet{mcinnes2017accelerated} for a more formal treatment of this topic.

Let $\vec{X} = \Set{\vec{x}_1, \dots, \vec{x}_{n}}$ be a dataset with $n$ feature vectors $\vec{x}_i$ and an associated distance metric $\dist{\vec{x}_{i}, \vec{x}_{j}}$. HDBSCAN* operates on mutual reachability distances, defined as~\citep{lelis2009mutual}:
\begin{equation}
 \label{eq:mutual-reachability}
 \mutreach{\vec{x}_{i}, \vec{x}_{j}} = \max*{\core{\vec{x}_{i}}, \core{\vec{x}_{j}}, \dist{\vec{x}_{i}, \vec{x}_{j}}},
\end{equation}
where $\core{\vec{x}_{i}}$ is the \emph{core distance} of $\vec{x}_{i}$, defined as the distance to its $k$-th nearest neighbour. HDBSCAN* converts these distances to densities as~\citep{campello2015hdbscan}:
\begin{equation}
 \label{eq:density-merge}
 \lambda(\vec{x}_i, \vec{x}_j) = \frac{1}{\mutreach{\vec{x}_{i}, \vec{x}_{j}}}.
\end{equation}
The mutual reachability distance effectively flattens all regions with a density higher than the local core distance. Consequently, higher $k$s produce smoother density profiles, reducing the number of local density maxima~\citep{campello2015hdbscan}. 

Single-linkage clustering on the mutual reachability distances efficiently recovers the clusters present at each density threshold~\citep{tarjan1975union-find,mcinnes2017accelerated}. The resulting dendrogram is pruned with a minimum cluster size threshold ($m_{\rm c}$) to form a cluster hierarchy that models the density contour tree~\citep{stuetzle2010pruning}. \citet{campello2015hdbscan} recommend treating $m_{\rm c} = k$ as a single tunable parameter, which is the default in implementations by \citet{mcinnes2017accelerated} and \citet{mcinnes2023fasthdbscan}.

\citet{campello2015hdbscan} propose two strategies to select clusters in the hierarchy. The \emph{leaf} strategy selects all local density maxima, i.e., all leaf-nodes. The \emph{excess of mass} (EOM) strategy prioritises nodes with longer \emph{cluster lifetimes}~\citep{fred2005lifetimes} in a larger region in feature space~\citep{muller1991eom}. \citet{malzer2020hybrid} limit the minimum density of selected EOM clusters, effectively blending DBSCAN with HDBSCAN*. All three strategies can select clusters at varying densities.

\section{Topological description}%
\label{sec:maths}
\begin{algorithm*}[t!]
  \caption{Our PLSCAN algorithm. The first steps are very similar to the \citet{mcinnes2017accelerated} algorithm for computing HDBSCAN*~\citep{campello2015hdbscan}. Later steps are specialised to support precomputed sparse distance matrices that form minimum spanning forests. The leaf tree and persistence trace form the core of PLSCAN's novel cluster selection strategy. Pseudocode for the later steps is available in Appendix~\ref{sec:appendix:plscan}.}%
  \label{alg:plscan}
  \begin{algorithmic}[1]
    \Function{PLSCAN}{$\vec{X}$, $k$, \texttt{sample\_weights}}
      \LComment{Computes PLSCAN clusters where $\vec{X}=\Set{\vec{x}_0, \dots,\vec{x}_n}$ is the data set, $k$ controls the number of nearest neighbors, and \texttt{sample\_weights} optionally describes how much each point contributes to a cluster's size.}
      \State \texttt{spanning\_tree} $\gets$ compute mutual reachability MST                                                                              \Comment{as in \citet{mcinnes2017accelerated}.}
      \State \texttt{linkage\_tree} $\gets$ compute single-linkage tree                                                                                   \Comment{as in \citet{mcinnes2017accelerated}.}
      \State \texttt{condensed\_tree} $\gets$ \Call{condense\_tree}{\texttt{linkage\_tree}, \texttt{spanning\_tree}, $n$, $k$}                            \Comment{using Alg.~\ref{alg:condense-tree}.}
      \State \texttt{leaf\_tree} $\gets$ \Call{leaf\_tree}{\texttt{condensed\_tree}, $n$, $k$}                                                            \Comment{using Alg.~\ref{alg:leaf-tree}.}
      \State \texttt{trace} $\gets$ \Call{persistence\_trace}{\texttt{leaf\_tree}, \texttt{condensed\_tree}, $n$}                                         \Comment{using Alg.~\ref{alg:persistence-trace}.}
      \State \texttt{labels}, \texttt{probabilities} $\gets$ \Call{select\_clusters}{\texttt{trace}, \texttt{leaf\_tree}, \texttt{condensed\_tree}, $n$}  \Comment{using Alg.~\ref{alg:select-clusters}.}
      \State \Return \texttt{labels}, \texttt{probabilities}
    \EndFunction
  \end{algorithmic}
\end{algorithm*}

Density-based clustering relates to concepts from algebraic topology. Topological data analyses describe datasets through global, qualitative features. For example, \emph{homology} measures the number of connected components, loops, and higher-dimensional voids~\citep[see][]{bartocci2013poincare}. 

In practice, these features are computed with \emph{simplicial complexes} that describe the connectivity between data points. The features that exist depend on the level of connectivity. For example, the connected components in a Vietoris--Rips complex containing edges when $\dist{\vec{x}_i, \vec{x}_j} < \e$ depend on the distance threshold $\e$~\citep[e.g.,][]{edelsbrunner2010topology}. Equivalently, DBSCAN clusters depend on the distance threshold. 

To avoid selecting a single threshold, \emph{persistent homology} describes topological features over all connectivity scales~\citep{edelsbrunner2002persistence}. Similar to the cluster lifetimes used by \citet{fred2005lifetimes} and \citet{leung2000scalespace}, the distance range at which features \emph{persist} is used to separate structural features from noisy features~\citep{edelsbrunner2010topology,carlsson2009topology}. 

PLSCAN applies the same principles, tracking local density maxima over all $m_{\rm c}$ thresholds and using their $m_{\rm c}$-persistence as a proxy for their relevance. Appendix~\ref{sec:appendix:tda} formalises these concepts, defining a metric space for which zero-dimensional persistent homology uncovers the minimum cluster size range in which leaf-clusters exist. This formulation highlights the generality of our approach, showing it can be applied to other monotonic linkage hierarchy pruning methods~\citep[e.g.,][]{stuetzle2010pruning, kpotufe2011pruning}.

Within this framework, our main contribution is an efficient algorithm for computing the leaf-cluster hierarchy from a single HDBSCAN* condensed tree constructed with an initial $m_{\rm c} \ge 2$. Section~\ref{sec:algorithm} provides a detailed description of this algorithm.

\section{Computing PLSCAN}%
\label{sec:algorithm}

This section describes PLSCAN from a computational perspective. Algorithm~\ref{alg:plscan} outlines the six main steps. The first three steps are very similar to HDBSCAN* and compute (mutual reachability) minimum spanning trees, linkage trees and condensed trees. We refer to \citet{campello2015hdbscan} and \citet{mcinnes2017hdbscan} for more details on these steps. The last three steps are novel. These steps compute the minimum cluster size leaf-cluster hierarchy and select persistent leaf-clusters (Sec.~\ref{sec:algorithm:leaf-tree}--\ref{sec:algorithm:cluster-selection}). 

The description starts at the \emph{leaf tree} construction step, which takes a condensed tree $\graph{C}$ as input: 
\begin{equation}
  \label{eq:condensed-tree}
  \graph{C} = \Set*{\left({r_{\rm parent}, r_{\rm child}, d_{\rm mut}, s}\right)},
\end{equation} 
where $r_{\rm parent}$ and $r_{\rm child}$ identify the connected components being merged, $d_{\rm mut}$ is the mutual reachability distance at which the merge occurs, and $s$ is the child's size. Our condensed tree is constructed in distance-sorted order (Alg.~\ref{alg:condense-tree}) to facilitate finding the distances at which clusters reach specific sizes. Additionally, we use a \emph{phantom root} that represents the whole dataset to work with precomputed minimum spanning forest inputs.

\subsection{Leaf tree}%
\label{sec:algorithm:leaf-tree}
\begin{figure*}[t]
  \centering
  \hfill{}%
  \begin{subfigure}[t]{0.32\textwidth}
    \centering
    Condensed trees
    \includegraphics[width=\textwidth]{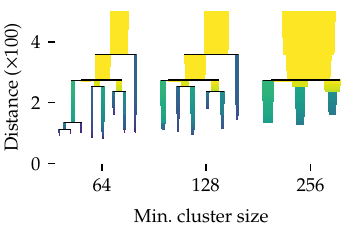}
    \caption{}%
    \label{fig:algorithm:leaf-tree:condensed-trees}
  \end{subfigure}%
  \hfill{}%
  \begin{subfigure}[t]{0.32\textwidth}
    \centering
    Annotated input
    \includegraphics[width=\textwidth]{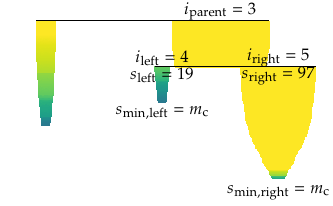}
    \caption{}%
    \label{fig:algorithm:leaf-tree:step-1}
  \end{subfigure}%
  \begin{subfigure}[t]{0.33\textwidth}
    \centering
    Leaf tree $\graph{F}$
    \begingroup{}
    \setlength{\tabcolsep}{4pt}
    \begin{tabular}{ccrrrr}
       & & & & & \\
      $i$ & $i_{\rm parent}$ & $d_{\rm min}$ & $d_{\rm max}$ & $s_{\rm min}$ & $s_{\rm max}$ \\
      $0$ & \cellcolor{red!30} $0$ & \cellcolor{red!30} $0.00$ & \cellcolor{red!30} $8.14$ & $25$ & \cellcolor{red!30} $150$ \\
      $1$ & \cellcolor{red!30} $0$ & $8.14$ & \cellcolor{red!30} $8.14$ & \cellcolor{violet!30} $25$ & $25$ \\
      $2$ & \cellcolor{violet!30} $1$ & $0.92$ & \cellcolor{violet!30} $8.14$ & \cellcolor{red!30} $m_{\rm c}$ & \cellcolor{violet!30} $25$ \\
      $3$ & \cellcolor{violet!30} $1$ & $1.74$ & \cellcolor{violet!30} $8.14$ & \cellcolor{teal!30} $19$ & \cellcolor{violet!30} $25$ \\
      $4$ & \cellcolor{teal!30} $3$ & $1.17$ & \cellcolor{teal!30} $1.74$ & \cellcolor{red!30} $m_{\rm c}$ & \cellcolor{teal!30} $19$ \\
      $5$ & \cellcolor{teal!30} $3$ & $0.48$ & \cellcolor{teal!30} $1.74$ & \cellcolor{red!30} $m_{\rm c}$ & \cellcolor{teal!30} $19$ \\
    \end{tabular}
    \endgroup{}
    \caption{}%
    \label{fig:algorithm:leaf-tree:output}
  \end{subfigure}%
  \caption{Leaf tree construction explainer. (\textbf{a}) Condensed trees at multiple minimum cluster sizes for the dataset from Fig.~\ref{fig:algorithm:smoothing}. The tree at the lowest threshold contains all merges present in the later trees. The size and distance at which segments merge remain constant regardless of the threshold. (\textbf{b}) A simple condensed tree $\graph{C}$ with two cluster merges and $\N = 150$, annotated for the first cluster merge. Cluster segments are identified by their leaf tree index $i$. (\textbf{c}) The resulting leaf tree, where $i$ indicates the leaf tree index that serves as an identifier. The background colours indicate when values were written: red for default values, teal for the first step, and violet for the second step. The uncoloured values are written in separate post-processing steps.}%
  \label{fig:algorithm:leaf-tree}
\end{figure*}

\begin{figure*}[t]
  \centering
  \begin{subfigure}{0.32\textwidth}
    \centering
    \includegraphics[width=\textwidth]{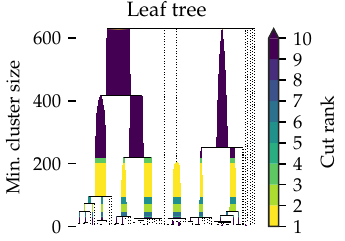}
    \caption{}%
    \label{fig:algorithm:plscan:leaf-tree}
  \end{subfigure}%
  \hfill{}%
  \begin{subfigure}{0.32\textwidth}
    \centering
    \includegraphics[width=\textwidth]{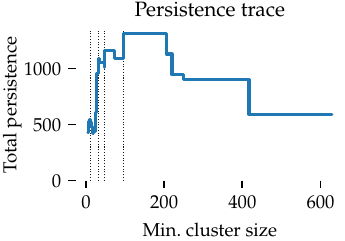}
    \caption{}%
    \label{fig:algorithm:plscan:persistence-trace}
  \end{subfigure}%
  \hfill{}%
  \begin{subfigure}{0.32\textwidth}
    \centering
    \includegraphics[width=\textwidth]{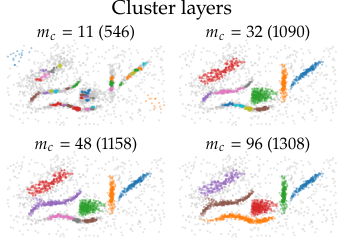}
    \caption{}%
    \label{fig:algorithm:plscan:cluster-layers}
  \end{subfigure}%
  \caption{Novel PLSCAN concepts demonstrated on the data from Fig.~\ref{fig:algorithm:smoothing}. (\textbf{a}) The leaf tree describes which local density maxima exist at each cluster size threshold. Colours indicate the top-10 highest total persistence peaks. Icicle widths encode the clusters' \emph{excess of mass}, i.e., the distance persistence sum over all points in the cluster~\citep{campello2015hdbscan}. The clusters' birth distances increase with the minimum cluster size (Fig.~\ref{fig:algorithm:leaf-tree:condensed-trees}), leading to smaller persistences at higher thresholds. (\textbf{b}) The persistence trace quantifies clustering `quality'. We compute the total size persistence over all leaf clusters that exist at a particular minimum cluster size threshold. Dotted lines indicate local persistence trace maxima. (\textbf{c}) Peaks in the persistence trace represent other stable clusterings. PLSCAN can efficiently compute flat clusterings for these peaks with its \emph{cluster layers}. The values between brackets indicate the total persistence for each layer.}%
  \label{fig:algorithm:plscan}
\end{figure*}

This step converts a condensed tree $\graph{C}$ into a \emph{leaf tree} by finding the cluster size ranges $(s_{\rm min}, s_{\rm max}]$ for which its cluster segments are leaf-clusters. Figure~\ref{fig:algorithm:leaf-tree:condensed-trees} demonstrates the intuition behind our approach by showing condensed trees computed at multiple minimum cluster sizes for the dataset from Fig~\ref{fig:algorithm:smoothing}. Notice that the tree at the lowest shown threshold ($m_{\rm c} = 64$) contains all merges present in the larger trees. The size ($s$) and distance ($d_{\rm mut}$) at which merges occur do not change with the $m_{\rm c}$ threshold. Instead, increasing $m_{\rm c}$ only prunes branches that do not reach the minimum size. Consequently, the sizes at which segments disappear or become leaves can be read directly from a single condensed tree constructed with a low $m_{\rm c}$. Figure~\ref{fig:algorithm:plscan:leaf-tree} shows the leaf tree for the cluster hierarchies from Fig.~\ref{fig:algorithm:leaf-tree:condensed-trees}. Read from left to right, this example shows the effect of progressively stronger pruning: existing merge heights stay fixed, while short-lived icicles disappear and expose their parents as new leaves. The leaf tree records exactly those minimum cluster size intervals.

Leaf trees are stored as a list of cluster segments:
\begin{equation}
  \label{eq:leaf-tree}
  \graph{F} = \Set*{\left({i_{\rm parent}, d_{\rm min}, d_{\rm max}, s_{\rm min}, s_{\rm max}}\right)},
\end{equation} 
where $i_{\rm parent}$ is the parent's leaf tree index and $[d_{\rm min}, d_{\rm max})$ is the mutual reachability distances range for which cluster segments exist.

The construction procedure (Alg.~\ref{alg:leaf-tree}) iterates through the condensed tree $\graph{C}$ in increasing distance order. This order guarantees children are processed before their parents, starting with the densest leaves. Each iteration processes a merge between two cluster segments. Such merges form row-pairs in $\graph{C}$ with the same $r_{\rm parent}$ and $d_{\rm mut}$ but different $r_{\rm child}$ and $s$ values.

In each iteration, both children fill in their maximum distance $d_{\rm max} = d_{\rm mut}$ and parent identifier $i_{\rm parent} = r_{\rm parent} - \N$. The children also set their maximum $m_{\rm c}$ thresholds:
\begin{equation}
  \label{eq:leaf-tree:death-size}
  s_{\rm max} = \min*{s_{\rm left}, s_{\rm right}},
\end{equation} 
as both children stop being leaves when either stops existing. The parent sets its minimum $m_{\rm c}$ threshold as:
\begin{equation}
  \label{eq:leaf-tree:birth-size}
  s_{\rm min} = \max*{s_{\rm max,left}, s_{\rm min,left}, s_{\rm min,right}}.
\end{equation}
This update rule delays the parent's birth until all downstream leaves have disappeared. Note, when Eq.~\ref{eq:leaf-tree:birth-size} is evaluated, $s_{\rm max,left} = s_{\rm max,right}$ as both children set the same maximum size in Eq.~\ref{eq:leaf-tree:death-size}. Figures~\ref{fig:algorithm:leaf-tree:step-1} and \ref{fig:algorithm:leaf-tree:output} demonstrate an iteration on a simple condensed tree with two cluster merges and $\N = 150$.

Afterwards, leaf tree segments with $s_{\rm min} < s_{\rm max}$ are leaf-clusters for the range $(s_{\rm min}, s_{\rm max}]$. Segments with $s_{\rm min} \geq s_{\rm max}$ do not become leaf-clusters for any $m_{\rm c}$ value and can be ignored in later steps. The minimum distances $d_{\rm min}$ and (phantom) root size limits are updated in a post-processing step. 

\subsection{Persistence trace}%
\label{sec:algorithm:persistence-trace}

The next step prepares the information needed to extract clusters from the leaf tree. This step computes persistence scores for all leaf-clusters at a $m_{\rm c}$. Specifically, we define a leaf cluster's \emph{size} persistence as:
\begin{equation}
  \label{eq:size-persistence}
  p_{\rm size} = s_{\rm max} - s_{\rm min}.
\end{equation}
The sum of these scores is used as a quality measure for the clusters at that $m_{\rm c}$, effectively rating a single cut through the leaf tree.

To select an $m_{\rm c}$ value, we need to rate all possible cuts through the leaf tree. This is possible because peaks in the total persistence can only occur when leaves merge or disappear, which happens at the $s_{\rm min}$ and $s_{\rm max}$ values in the leaf tree. Between two consecutive critical values, the set of active leaf-clusters does not change. Consequently, the total size persistence is constant on that interval. We therefore only need to compute the total persistence at these discrete $m_{\rm c}$ values. Figure~\ref{fig:algorithm:plscan:persistence-trace} shows the resulting \emph{persistence trace}, indicating the total persistence at all $m_{\rm c}$ cuts for the leaf tree from Fig.~\ref{fig:algorithm:plscan:leaf-tree}. The pseudocode for our persistence trace construction procedure is available in Alg.~\ref{alg:persistence-trace}. 

\subsection{Cluster selection}%
\label{sec:algorithm:cluster-selection}
\begin{figure*}[!t]
  \centering
  \includegraphics[width=\textwidth]{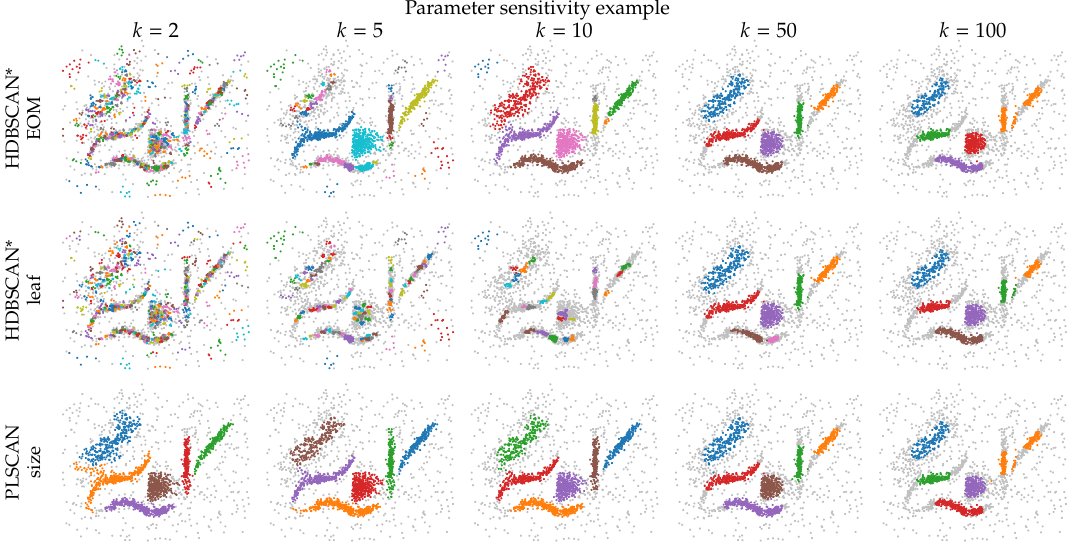}%
  \caption{Clusterings for different values of $k$ on the data from Fig.~\ref{fig:algorithm:smoothing}. PLSCAN automatically finds a `good' minimum cluster size threshold. Consequently, its clusterings vary predictably with $k$. For example, the detected clusters shrink at higher $k$ values. HDBSCAN*'s clusterings, on the other hand, vary considerably with $k$ when the minimum cluster size is also set to $k$~\citep[which is the default in][]{mcinnes2017hdbscan,mcinnes2023fasthdbscan}.}%
  \label{fig:algorithm:plscan:sensitivity}
\end{figure*}

The final step computes cluster labels and membership probabilities (Alg.~\ref{alg:select-clusters}). By default, PLSCAN selects the $m_{\rm c}$ value with the highest total persistence, extracting the corresponding leaf-clusters from the leaf tree. Additionally, analysts can inspect clusterings at other local persistence maxima to explore alternative segmentations. Figure~\ref{fig:algorithm:plscan:cluster-layers} shows segmentations at the four strongest persistence peaks from Fig.~\ref{fig:algorithm:plscan:persistence-trace}. 

\subsection{Parameter sensitivity}%
\label{sec:algorithm:sensitivity}

PLSCAN's automatic $m_{\rm c}$ selection reduces its sensitivity to the number of neighbours $k$. For example, Fig.~\ref{fig:algorithm:plscan:sensitivity} compares HDBSCAN* (with $m_{\rm c} = k$) and PLSCAN clusterings for different $k$ values on the dataset from Fig.~\ref{fig:algorithm:smoothing}. HDBSCAN*'s labels vary considerably at low $k$s. The leaf cluster selection strategy, in particular, requires large $k$ values to find the expected clusters. In contrast, PLSCAN finds similar clusterings for all tested $k$s, with the detected clusters shrinking as $k$ increases. Section~\ref{sec:demo:sensitivity} explores whether this pattern generalises to real-world datasets.

\subsection{Stability}%
\label{sec:algorithm:stability}

Like HDBSCAN*, the PLSCAN algorithm relies on deterministic density-based clustering principles. Consequently, PLSCAN provides similar clustering results when repeatedly evaluated on (different) samples of an underlying distribution. In contrast, $k$-Means uses random initialisations~\citep{lloyd1982kmeans} and genetic clustering algorithms use stochastic optimisation~\citep{xu2005survey,ezugwu2022survey}, producing different results on each run. The $k$-Means++ intialisation strategy by \citet{arthur2006kmeans} reduces this instability for $k$-Means by spreading out the initial centroids. Section~\ref{sec:demo:stability} evaluates the stability of PLSCAN in practice.

\subsection{Computational complexity}%
\label{sec:algorithm:complexity}
\begin{table*}[t]
  \centering
  \caption{Datasets used in the case studies with their pre-processing strategy. Small cosine-distance datasets were L2~normalised. Large image and text datasets were UMAP-projected~\citep{mcinnes2018umap} to $50$ dimensions using cosine distances, $15$ neighbours, and increasing repulsion strengths ($0.001$, $0.01$, $0.1$, $1$). CellCycle-237 was z-score normalised~\citep[as in][]{castro2019unified}. The remaining numerical datasets were clustered on original features. CIFAR-10 and 20~Newsgroups were converted to a numerical space using pre-trained CLIP~\citep{radford2021clip} and MiniLM~\citep{reimers2019sentence,wang2020minilm} models, respectively, before their UMAP dimensionality reduction step. The C.~elegans dataset was pre-processed as in \citet{packer2019elegans} before its L2-normalisation. All algorithms were evaluated on identical inputs using Euclidean distances.}%
  \label{tab:demo:sensitivity:datasets}
  \begin{tabular}{lrrrr}
    \textbf{Dataset} & \textbf{Num.~points} & \textbf{Num.~dimensions} & \textbf{Num.~classes} & \textbf{Processing} \\
    Articles-1442-5~\citep{naldi2011articles}    &    253 & 4,636 &  5  & L2~norm. \\
    Articles-1442-80~\citep{naldi2011articles}   &    253 &   388 &  5  & L2~norm. \\
    Audioset (music)~\citep{gemmeke2017audioset} & 26,629 & 1,280 & 51  & UMAP    \\
    Authorship~\citep{simonoff2003analyzing}     &    841 &    70 &  4  & L2~norm. \\
    CTG~\citep{ayres2000sisporto}                &  2,126 &    35 & 10  & L2~norm. \\
    CellCycle-237~\citep{yeung2001model}         &    384 &    17 &  5  & Z-score \\
    CIFAR-10~\citep{krizhevsky2009learning}      & 50,000 &   784 & 10  & UMAP    \\
    E.~coli~\citep{horton1996probabilistic}      &    336 &     7 &  7  &         \\
    C.~elegans~\citep{packer2019elegans}         &  6,188 &    50 & 26  & L2~norm. \\
    Fashion MNIST~\citep{xiao2017fashion}        & 70,000 &   784 &  9  & UMAP    \\
    Iris~\citep{unwin2021iris}                   &    150 &     4 &  2  &         \\
    MFeat-factors~\citep{jain2000statistical}    &  2,000 &   216 & 10  &         \\
    MFeat-Karhunen~\citep{jain2000statistical}   &  2,000 &    64 & 10  &         \\
    MNIST~\citep{lecun1998mnist}                 & 70,000 &   784 &  9  & UMAP    \\
    20 Newsgroups~\citep{lang1995newsweeder}     & 11,314 & 5,000 & 19  & UMAP    \\
    Semeion Digits~\citep{buscema1998metanet}    &  1,593 &   256 & 10  & L2~norm. \\
    Yeast Galactose~\citep{yeung2003clustering}  &   205  &   80  &  4  &         \\
  \end{tabular}
\end{table*}

Like HDBSCAN*~\citep{mcinnes2017accelerated}, PLSCAN's computational cost is dominated by the mutual reachability minimum spanning tree construction. The space-tree accelerated nearest neighbour queries have an average time complexity of $O(\N \log \N)$ on low-dimensional datasets~\citep{bentley1975kdtree,omohundro1989five}. At higher dimensions, these steps can degrade to $O(\N^2)$~\citep{weber1998quantitative}. \boruvka{}'s algorithm has a time complexity of $O(\N \log^2 \N)$~\citep{nesetril2001boruvka}, with the same high-dimensional data caveat, as our implementation utilises space trees to find nearest neighbours in each iteration.

The other steps are listed here for completeness.
\begin{itemize}
  \item Computing the linkage tree has a time complexity of $O(\N \alpha(\N))$, where $\alpha$ is the inverse Ackermann function~\citep{tarjan1975union-find}. 
  \item The condensed tree computations are in $O(\N)$, reducing to one iteration over the linkage tree with at most $\N - 1$ edges. 
  \item The leaf tree computation iterates over the condensed tree, which contains all points $\N$ and has at most ${\rm L} = (\N - 1) / m_{\rm c}$ cluster segments. Therefore, the worst-case complexity is $O(\N^2)$. In practice, complexities closer to $O(\N)$ are observed because usually ${\rm L} \ll \N$. 
  \item The size persistence trace computation follows a similar pattern, with a time complexity of $O(\rm L)$, where $\rm L$ is the number of cluster segments in the condensed tree and leaf tree. 
  \item The other persistence measures propagate all data points up the leaf tree hierarchy, resulting in a higher time complexity of $O({\rm L} \N)$. 
  \item Cluster selection is in $O({\rm L} + \N)$, iterating over the leaf and condensed trees to assign cluster labels. 
\end{itemize}
Section~\ref{sec:demo:cost} demonstrates computational costs in practice.

\section{Case studies}%
\label{sec:demo}

This section evaluates PLSCAN on real-world datasets and compares its parameter sensitivity, stability, and computational costs with those of HDBSCAN*.

All experiments were run on Python 3.14 with the \texttt{hdbscan}~\citep[version 0.8.43,][]{mcinnes2017hdbscan}, \texttt{fast\_hdbscan}~\citep[version 0.3.2,][]{mcinnes2023fasthdbscan}, \texttt{scikit-learn}~\citep[version 1.8.0,][]{pedregosa2011scikit}, and \texttt{plscan} (version 0.2.0) packages. The test machine was equipped with an 8-core CPU and 32 GB of RAM.

\subsection{Parameter sensitivity benchmark}%
\label{sec:demo:sensitivity}
\begin{figure*}[t]
  \centering
  \hfill{}%
  \begin{subfigure}{0.49\textwidth}
    \centering
    \includegraphics[width=\textwidth]{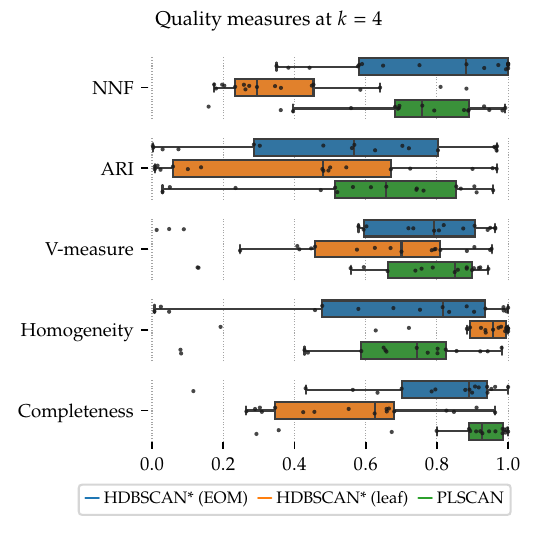}
    \caption{}%
    \label{fig:demo:sensitivity:boxplots:measures}
  \end{subfigure}%
  \hfill{}%
  \begin{subfigure}{0.49\textwidth}
    \centering
    \includegraphics[width=\textwidth]{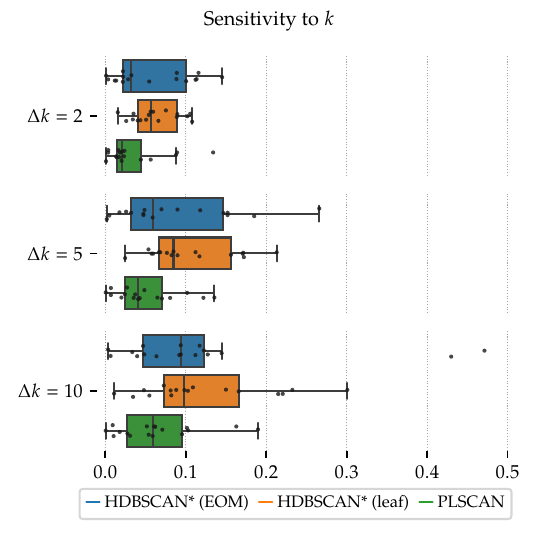}
    \caption{}%
    \label{fig:demo:sensitivity:boxplots:perturbations}
  \end{subfigure}%
  \hfill{}%
  \caption{Performance measure and $k$-sensitivity distributions for each algorithm configuration. \textbf{(a)} The evaluated clustering performance measures over all datasets at $k=4$, summarising \emph{default} performances following the parameter configuration by \citet{campello2015hdbscan} in their evaluations of HDBSCAN*. \textbf{(b)} The $\Delta k$-sensitivity distributions for each algorithm configuration. These distributions summarise the per-dataset ARI--$k$ curves shown in Fig.~\ref{fig:demo:sensitivity:curves}. \citet{peng2022cdc} classify scores $\le 0.25$ as \emph{insensitive}.}%
  \label{fig:demo:sensitivity:boxplots}
\end{figure*}

PLSCAN was less sensitive to the number of neighbours $k$ than HDBSCAN* on the example dataset from Fig.~\ref{fig:algorithm:plscan:sensitivity}. More specifically, we observed that PLSCAN found the same clusters regardless of the tested $k$-values, while HDBSCAN* only found the expected clusters at higher $k$ values. 

In this section, we perform a sensitivity analysis to explore whether that pattern generalises to other real-world datasets, following a methodology previously used for this purpose by \citet{peng2022cdc}. We aim to determine how the clustering performance of PLSCAN compares to HDBSCAN* at default and optimised parameter values, and how sensitive both algorithms are to changes in $k$.

\paragraph*{Datasets} 
Several real-world datasets were used in this benchmark. Table~\ref{tab:demo:sensitivity:datasets} lists the number of observations and features of these datasets. Most of the datasets were previously used to evaluate clustering algorithms by \citet{campello2015hdbscan} or \citet{castro2019unified}. The \emph{CIFAR-10} and \emph{20 Newsgroups} datasets required conversion to a numerical feature space. We applied pre-trained CLIP~\citep{radford2021clip} and MiniLM~\citep{reimers2019sentence,wang2020minilm} models, respectively.

The \emph{Audioset (music)} dataset is a manually created subset of the \emph{Audioset} dataset. We selected the leaf nodes under \emph{Music genre} in the Audioset ontology~\citep{gemmeke2017audioset} to describe the most specific genres. Then, we selected the observations with exactly one of those labels from the unbalanced training set to create a multi-class dataset. The final dataset contains $26,629$ samples with $51$ classes and $1,280$ features computed by \citet{hershey2017cnn}.

Datasets were preprocessed according to their type. Small datasets for which the cosine distance is appropriate were L2~normalised and clustered with the Euclidean distance to approximate the cosine measure because neither HDBSCAN* nor PLSCAN support cosine distances in their space-tree accelerated MST computations. Large datasets with image or text embeddings were projected to $50$ dimensions using UMAP~\citep{mcinnes2018umap} with cosine distances, $15$ neighbours, and increasing repulsion strengths ($0.001$, $0.01$, $0.1$, $1$) to improve layout convergence. The resulting projections were then clustered with Euclidean distances. Reducing the number of dimensions avoids large computation costs for the clustering step (see Sec.~\ref{sec:demo:cost}). CellCycle-237 was z-score normalised and clustered with Euclidean distances, following \citet{castro2019unified}. The remaining small numerical datasets were clustered without preprocessing using Euclidean distances. Because all algorithms were evaluated on the same preprocessed inputs, the benchmark compares their relative clustering performance.

\paragraph*{Experimental setup} 
\begin{figure*}[t]
  \centering
  \includegraphics[width=\textwidth]{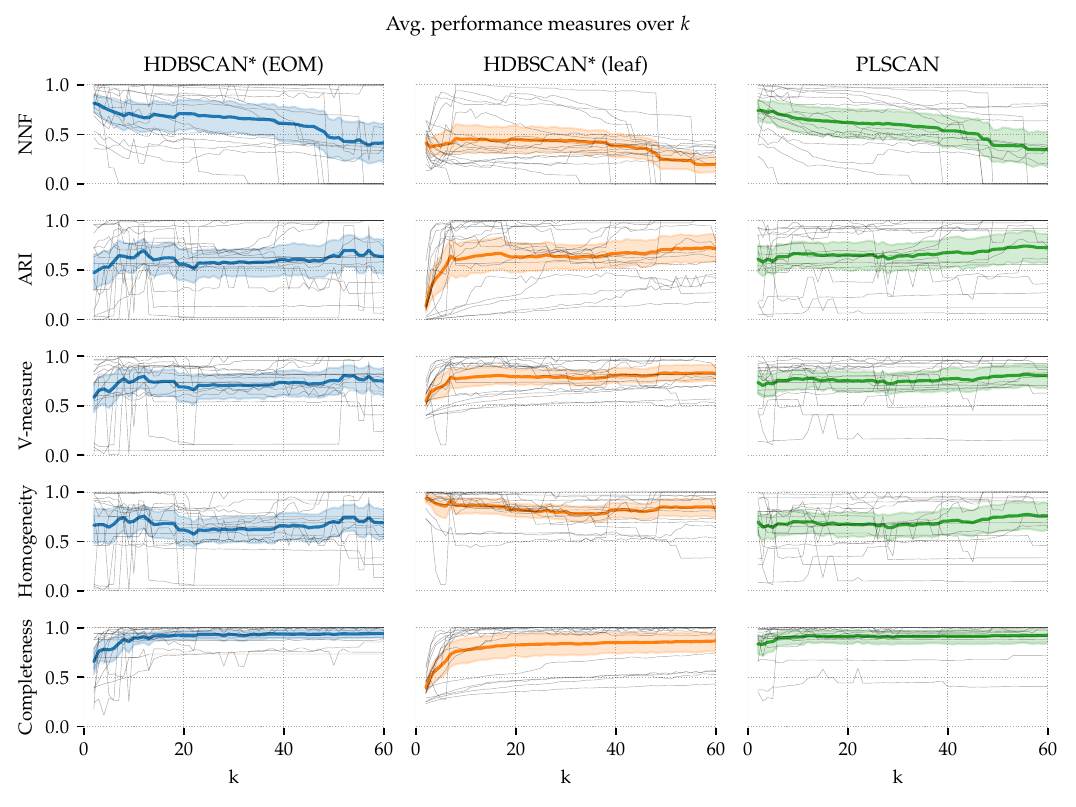}
  \caption{The average performance measures (rows) across all datasets over $k$, coloured by algorithm configuration (columns). Shaded areas indicate the $95\%$ confidence intervals of the means. Smaller black lines encode the performances for the individual datasets in Tab.~\ref{tab:demo:sensitivity:datasets}.}%
  \label{fig:demo:sensitivity:curves}
\end{figure*}

We measure sensitivity to the main parameter $k$ using \emph{adjusted rand index} (ARI) scores~\citep{hubert1985comparing}, following a methodology by \citet{peng2022cdc} based on \emph{Latin-hypercube one-factor-at-a-time}~\citep{griensven2006sensitivity}. First, we constrain $k$ to the range $[2, 50]$ and divide that range into $\N=10$ evenly spaced segments. Then, we sample a value $k_{i}$ from each segment, combining randomness with a uniform coverage of the parameter space. Each sampled $k_{i}$ gets perturbed by a small value $\pm \Delta k \in \Set*{2, 5, 10}$, randomly selecting the positive or negative direction. Finally, we compute the sensitivity at a given perturbation $\Delta k$ as:
\begin{equation}
\label{eq:sensitivity}
s_{\Delta k} = \frac{1}{\N} \sum_{i=1}^{\N} \left| \frac{\ari{k_{i} \pm \Delta k} - \ari{k_{i}}}{\ari{k_{i} \pm \Delta k} + \ari{k_{i}}} \right|,
\end{equation}
where $\ari{k}$ computes the ARI at a particular $k$ value. The sampling procedure was repeated $15$ times, and Eq.~\ref{eq:sensitivity} was averaged over those repetitions. Each dataset and algorithm was evaluated on the same sampled $k_{i}$ values to ensure a fair comparison between methods. 

Aside from the ARI scores, three additional diagnostic clustering performance measures were also collected for each dataset and algorithm. Namely, we measured whether clusters mostly contained points from a single class (Homogeneity), whether all points from a class were assigned to the same cluster (Completeness), and their combination (V-measure)~\citep{rosenberg2007v}. All performance measures were computed over only those points that were assigned to a non-noise cluster by the algorithms. We, therefore, also report the fraction of points not classified as noise as the \emph{non-noise fraction} (NNF).

Combined, this benchmark provides three complementary views of the same evaluations: performances at default parameter values (Fig.~\ref{fig:demo:sensitivity:boxplots:measures}), average performances measures over $k$ (Fig.~\ref{fig:demo:sensitivity:curves}), and the aggregated perturbation sensitivity derived from per-$k$ ARI scores (Fig.~\ref{fig:demo:sensitivity:boxplots:perturbations}). We include HDBSCAN*~\citep{mcinnes2017hdbscan} with the EOM and leaf cluster selection strategies and PLSCAN in this benchmark. Both algorithms set the (initial) minimum cluster size $m_{\rm c} = k$, which is their default behaviour. 

\paragraph*{Results} 
\begin{figure*}[t]
  \centering
  \includegraphics[width=.65\textwidth]{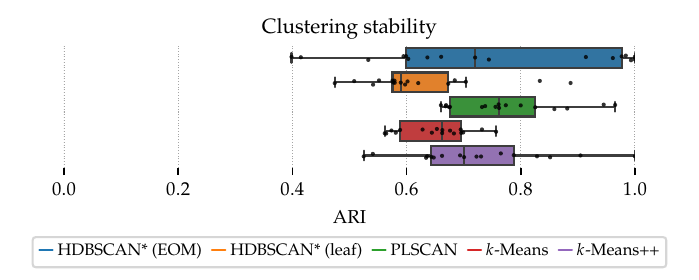}
  \caption{Distributions of clustering stability scores over the pre-processed dataset from Tab.~\ref{tab:demo:sensitivity:datasets}. Clustering stability was measured as the average ARI score over $15$ sub-sample pairs, inspired by \citet{ben2001stability}. Sub-samples were computed without replacement and contained 80\% of the data points.}%
  \label{fig:demo:stability}
\end{figure*}

Figure~\ref{fig:demo:sensitivity:boxplots:measures} visualises the algorithms' performance measure distributions at $k = 4$, following the parameter configuration by \citet{campello2015hdbscan} in their evaluations of HDBSCAN*. 
These results are reported as the algorithms' default performance, for cases where $k$ is not tuned or a fixed minimum cluster size is used. Note that $k = 4$ and our pre-processing may not be optimal, depending on the dataset. However, the resulting values do provide a performance indication for lower $k$ values. 

In this case, comparing PLSCAN to HDBSCAN* with EOM, it achieved higher median ARI ($0.66$ to $0.57$) and V-measure ($0.85$ to $0.79$) values, while giving more points the noise label ($0.76$ to $0.88$ NNF). This higher performance is achieved through higher median Completeness ($0.93$ to $0.89$) but lower median Homogeneity scores ($0.74$ to $0.82$). Consequently, the clusters from PLSCAN generally contained a relatively larger portion of a ground truth class but also relatively more points that did not belong to that class while maintaining a higher overall ARI.

HDBSCAN* with the leaf cluster selection strategy performed worse than the other two algorithms on all measures except for homogeneity, where it had the highest median score. This pattern is consistent with the leaf strategy's tendency to produce smaller clusters with more noise points at low $m_{\rm c}$ values~\citep{campello2015hdbscan}.

Figure~\ref{fig:demo:sensitivity:curves} illustrates how the performance measures changed with respect to $k$. Specifically, the plots show coloured curves for the average performance measures with 95\% confidence intervals, enabling a visual comparison of the algorithms' performance at different $m_{\rm c} = k$ values. In addition, fainter black lines show curves for the individual datasets, over which the averages were computed. The main patterns visible are:
\begin{itemize}
  \item NNF decreases with $k$ for all algorithms, indicating that more points are classified as noise at higher $k$ values.
  \item HDBSCAN* with the leaf cluster selection strategy has a steep drop in performance at low $k$ values, while the other algorithms remain more stable. 
  \item A similar drop in performance measures is visible for HDBSCAN* with EOM, although less pronounced. 
  \item PLSCAN did not show a clear drop in performance at low $k$ values, reproducing the pattern we found in Fig.~\ref{fig:algorithm:plscan:sensitivity}.
  \item PLSCAN has smoother average performance curves than HDBSCAN* with EOM, especially at $k < 20$.
\end{itemize}
Appendix~\ref{sec:appendix:performance} visualises per-dataset ARI--NNF curves over $k$, providing a more detailed view of the performance variations.

The actual sensitivity scores from Eq.~\ref{eq:sensitivity} are summarised in Fig.~\ref{fig:demo:sensitivity:boxplots:perturbations}. The figure displays the sensitivity score distributions over all datasets for each algorithm and perturbation size $\Delta k$. The results confirm that PLSCAN was less sensitive to changes in $k$ than HDBSCAN* with EOM, achieving lower median sensitivity scores $0.02$ to $0.03$ at $\Delta k=2$, $0.04$ to $0.06$ at $\Delta k=5$, and $0.06$ to $0.09$ at $\Delta k=10$. The sensitivity differences appear fairly stable across the perturbation sizes $\Delta k$, however \citet{peng2022cdc} classify scores $\le 0.25$ as \emph{insensitive}, which all algorithms achieved at all perturbation sizes.

\subsection{Stability benchmark}%
\label{sec:demo:stability}

Now, we turn to comparing PLSCAN's stability to HDBSCAN* and $k$-Means.
Inspired by \citet{ben2001stability}, we quantify stability as the average similarity between clusterings obtained from different sub-samples. This approach provides a practical measure of how robust the clustering results are to variations in the data.

\paragraph*{Datasets} 
This benchmark re-uses the pre-processed datasets from the sensitivity benchmark, shown in Tab.~\ref{tab:demo:sensitivity:datasets}.

\paragraph*{Experimental setup}
Each algorithm was evaluated on $15$ pairs of sub-samples for each dataset, where each sub-sample contained $80\%$ of the original dataset and was computed without replacement. The clustering similarity was computed as ARI scores over the points in both sub-samples. PLSCAN and HDBSCAN* were evaluated at a fixed $k = 4$, following the parameter configuration by \citet{campello2015hdbscan}. $k$-Means was evaluated with random initialisation and the $k$-Means++ initialisation strategy~\citep{arthur2006kmeans}. Both $k$-Means variants were given the true number of classes as input, which is infeasible in practice and provides a best-case scenario for their stability.

\paragraph*{Results} 
\begin{figure*}[t]
  \centering
  \includegraphics[width=\textwidth]{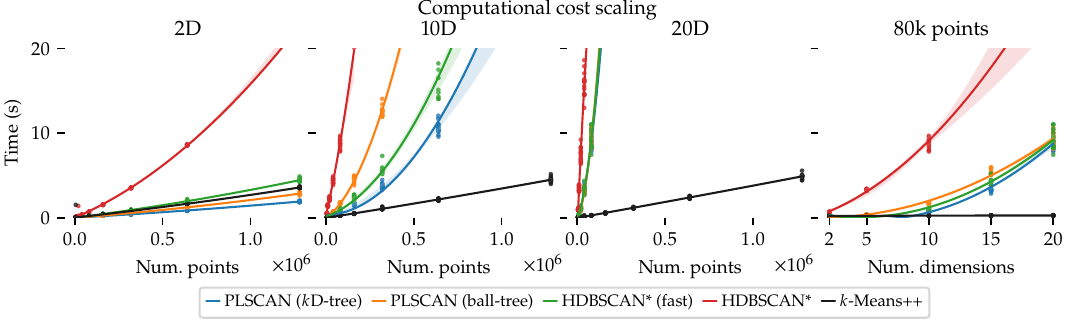}
  \caption{Quadratic regression lines relating computational cost to the number of data points and the number of dimensions. The left three plots show scaling over the number of data points in $2$, $10$, and $20$ dimensions. The right-most plot shows scaling over the number of dimensions at 80k data points. In this benchmark, the PLSCAN implementations (blue and orange) scale slightly better than \texttt{fast\_hdbscan}~\citep{mcinnes2023fasthdbscan} (green), which also uses a parallel $k$D-tree implementation. The $k$-Means++ algorithm (black) does not rely on space trees and scales independently of the number of dimensions.}%
  \label{fig:demo:cost}
\end{figure*}
Figure~\ref{fig:demo:stability} shows the resulting stability score distributions for each algorithm and dataset. PLSCAN achieved a higher median stability score ($0.76$) than HDBSCAN* with EOM ($0.72$) and $k$-Means++ ($0.70$), while HDBSCAN* with the leaf cluster selection strategy had the lowest median score ($0.59$) followed by $k$-Means with random initialisations ($0.66$). The stability difference between PLSCAN and HDBSCAN* with EOM ($0.76$ to $0.72$) was similar to the difference between $k$-Means++ and $k$-Means with random initialisations ($0.70$ to $0.66$).

\subsection{Computational cost benchmark}%
\label{sec:demo:cost}

Next, we compare the computational cost scaling of HDBSCAN*, PLSCAN, and $k$-Means++ over dataset size and dimensionality. Specifically, we compare the \texttt{hdbscan}~\citep{mcinnes2017hdbscan}, \texttt{fast\_hdbscan}~\citep{mcinnes2023fasthdbscan} and \texttt{scikit-learn}~\citep{pedregosa2011scikit} implementations and limit our discussion to general scaling trends~\citep[see][]{kriegel2017black}.

\paragraph*{Datasets}
This benchmark generated synthetic datasets with $100$ evenly sized clusters sampled from Gaussian distributions. The number of data points ($\N$) and the number of dimensions ($\D$) were varied. $15$ datasets were generated for each configuration to average out randomness in the run times.

\paragraph*{Experimental setup}
The algorithms were evaluated with mostly their default parameters. The number of neighbours was set to $k=10$, and the minimum cluster size was set to $m_{\rm c} = \min*{\N / 100, 100}$ for HDBSCAN* and PLSCAN. $k$-Means++ was given the ground-truth number of clusters as input. With these settings, \texttt{hdbscan} and \texttt{fast\_hdbscan} use a $k$D-tree to accelerate the main computational bottleneck: computing the mutual reachability minimum spanning tree. The latter implementation uses parallel nearest neighbour queries for further speed-ups, supporting only Euclidean distances. PLSCAN employs a similar parallel implementation that supports multiple distance metrics and ball-tree acceleration. Both the $k$D-tree and ball-tree PLSCAN implementations were evaluated in this benchmark.

All algorithms were given a timeout of \qty{20}{\second}. Algorithms exceeding that limit were not evaluated on datasets with the same size or larger. This timeout was applied to each number of dimensions separately. 

\paragraph*{Results} 
Figure~\ref{fig:demo:cost} displays the quadratic regression lines for the resulting computational costs, with shaded areas indicating the 95\% confidence intervals. The left three plots show curves at 2, 10, and 20 dimensions for increasing dataset sizes. The right-most plot shows curves at $80,000$ data points for increasing numbers of dimensions. 

As expected, all HDBSCAN* and PLSCAN implementations scale worse at higher dimensions, due to their reliance on space trees~\citep{weber1998quantitative}. The $k$-Means++ algorithm remains unaffected by the number of dimensions, due to its fundamentally different approach. Interestingly, at 2D, the $k$D-tree PLSCAN implementation was consistently faster than $k$-Means++. Additionally, our PLSCAN implementation was faster than \texttt{fast\_hdbscan}, likely due to additional parallelism and more efficient cluster selection and label assignment.

\section{Discussion}%
\label{sec:discussion}

The present paper proposes PLSCAN, a novel density-based clustering method intended for exploratory data analyses where few assumptions about the data are available. PLSCAN adapts HDBSCAN*'s linkage hierarchy construction with a new multiscale minimum cluster size filtration and persistence-based cluster selection procedure. The algorithm constructs an HDBSCAN* leaf-cluster hierarchy, listing all minimum cluster size thresholds at which local density maxima change. Appendix~\ref{sec:appendix:tda} demonstrates that this is equivalent to computing persistent homology using a novel distance metric. The algorithm provides a strategy for selecting stable clusterings at multiple levels of detail by detecting peaks in the total minimum cluster size persistence.

PLSCAN's performance was evaluated on several real-world datasets, achieving a higher median ARI than HDBSCAN* with EOM at $k = 4$ (Fig.~\ref{fig:demo:sensitivity:boxplots:measures}). A parameter sensitivity benchmark demonstrated PLSCAN's robustness to changes in $k$ (Fig.~\ref{fig:demo:sensitivity:boxplots:perturbations}), while the corresponding ARI--$k$ curves show how those differences evolve over the tested parameter range (Fig.~\ref{fig:demo:sensitivity:curves}). The analysis reproduced the pattern observed in a simple 2D dataset (Fig.~\ref{fig:algorithm:plscan:sensitivity}), where PLSCAN found similar clusters for all tested $k$ values, while HDBSCAN* only found the expected clusters at higher $k$ values. PLSCAN's median stability score exceeded HDBSCAN* with EOM by the same margin as $k$-Means++ exceeded $k$-Means with random initialisation (Fig.~\ref{fig:demo:stability}). A computational cost benchmark compared PLSCAN's run-times to HDBSCAN* and $k$-Means++ (Fig.~\ref{fig:demo:cost}). On low-dimensional datasets ($\D \le 5$, $\N \le 1.28\times 10^6$), our PLSCAN implementation was faster than scikit-learn's $k$-Means++ implementation with 100 clusters. At higher dimensions, PLSCAN's computational cost scales similarly to HDBSCAN*, while remaining faster than the implementation by \citet{mcinnes2023fasthdbscan} in this benchmark.

Overall, PLSCAN is suitable for exploratory analyses where expected cluster sizes are unknown, because the algorithm automatically identifies minimum cluster size thresholds that produce persistent clusters, removing a key parameter from the clustering process. By leveraging density-based clustering principles, PLSCAN provides descriptive results that identify persistent high-density regions, regardless of their shape. Additionally, ranking stable clusterings by their persistence enables analysts to explore multiple levels of detail. Whether density maxima are a good model for the ``true'' clusters researchers seek in their datasets with unknown structure depends on the use case~\citep{hennig2015trueclusters}. Consequently, while we evaluated a broad range of datasets, the applicability of PLSCAN can vary on a case-by-case basis.

\subsection{Using PLSCAN}%
\label{sec:discussion:practice}

In practice, we recommend using PLSCAN with $k=4$ as a starting point because low $k$ values produce less noise points (Fig.~\ref{fig:demo:sensitivity:curves}). \citet{ester1996dbscan} and \citet{campello2015hdbscan} similarly used $k=4$ in their evaluations of DBSCAN and HDBSCAN*, respectively. \citet{schubert2017dbscan} also recommend $k=4$ as a suitable default for DBSCAN on many datasets, but note that higher values may be beneficial for datasets with larger levels of noise. In PLSCAN, the decoupling of $k$ and $m_{\rm c}$ results in lower sensitivity and higher clustering performance on low $k$s, compared to HDBSCAN* (Sec.~\ref{sec:demo:sensitivity}).

It can be beneficial to inspect clusterings at multiple peaks in the persistence trace (Fig.~\ref{fig:algorithm:plscan:persistence-trace}). These peaks indicate stable clusterings at different levels of detail (Fig.~\ref{fig:algorithm:plscan:cluster-layers}). In this way, researchers can identify a managable number of reasonable clusterings and manually explore them to discover meaningful clusters. Visualising the complete cluster hierarchy can also provide insights about the relationships between clusters (Fig.~\ref{fig:algorithm:plscan:leaf-tree}).

\subsection{Clustering on embeddings}%
\label{sec:discussion:embeddings}

Extracting clusters from lower-dimensional projections can reduce computational costs (Sec.~\ref{sec:demo:cost}). However, a dimensionality reduction step introduces additional variability and uncertainty~\citep{chari2021dimreduc} and requires parameter tuning. This was not a problem in the parameter sensitivity benchmark because we were mainly interested in relative clustering differences between algorithms and $k$ values, not the absolute performances. In practice, researchers should ensure their projections accurately reflect the data's intrinsic structure.

\subsection{Limitations and future work}%
\label{sec:discussion:limitations}

The PLSCAN algorithm and the presented evaluation have several limitations and areas for future work. 

\paragraph*{Other persistence measures}

PLSCAN, as presented, only considers clusters' persistence over the minimum cluster size threshold when selecting clusters. However, the leaf tree structure also allows for computing persistence over other parameters. 

For example, our implementation can also compute distance $d$ and density \emph{$\lambda$} persistence measures. These measures complement size persistence by scoring how long a leaf-cluster remains separated in mutual reachability distance or density, rather than only over minimum cluster size thresholds. Note, $d_{\rm min}$ in the leaf tree (Eq.~\ref{eq:leaf-tree}) does not match the birth distance of a leaf cluster because they contain points from their child segments. Instead, we find the lowest distance point for which the leaf cluster exceeds $m_{\rm c}$. In contrast to \citet{campello2015hdbscan} in Eq.~\ref{eq:density-merge}, we constrain $\lambda$ to the range $[0, 1]$ by using:
\begin{equation}
  \label{eq:density}
  \lambda(\vec{x}_i, \vec{x}_j) = e^{-\mutreach{\vec{x}_{i}, \vec{x}_{j}}}.
\end{equation}

These alternatives can also be combined with the size persistence measure, forming \emph{size--$d$} and \emph{size--$\lambda$} bi-persistence measures. These bi-persistences favour leaf-clusters that remain separated over a broad distance or density range and do so across many $m_{\rm c}$ thresholds. Equivalently, they sum the $d$ and $\lambda$ persistences over all $m_{\rm c}$ values for which a leaf-cluster exists, so bi-persistences can be interpreted as an area in a size--distance or size--density plane.

Future research could explore the utility of these alternative persistence measures. For example, they may be more suitable for datasets with large size differences between clusters, where size persistence may not be informative. However, they also require additional computational costs and may be less interpretable than size persistence alone.

\paragraph*{Straight size cuts}
Selecting clusters by straight minimum cluster size cuts through the leaf tree may not be suitable for all datasets. Datasets with large size differences between clusters---such as the E.~coli dataset---may not be well represented by straight cuts. Alternatively, strong two-sided structures---where clusters combine into two sides before merging into a single connected component---may dominate the persistence trace, preventing smaller clusters from being selected by default.

Future work could explore multi-level cuts through the leaf tree to address this limitation. Similar techniques exist for hierarchical clustering~\citep[e.g.,][]{langfelder2008defining}. The main challenge is handling the non-binary hierarchy and conditionally not selecting any nodes in some branches. Otherwise, many small leaf clusters directly descending from the root would get selected.

\paragraph*{Multi-component inputs}
Inspired by \citet{jackson2018scaling}, PLSCAN supports precomputed minimum spanning forests and multi-component sparse distance matrices as inputs. In these cases, clusters are extracted from each connected component while optimising the total persistence over all components. This feature enables efficient cluster extraction from approximate nearest neighbour graphs, thereby avoiding the computational costs associated with minimum spanning tree extraction on high-dimensional datasets. 

Future work could formulate guidelines for clustering on approximate nearest neighbour graphs. Connected components in such graphs should contain at least two target clusters because PLSCAN cannot select the components themselves as clusters. Consequently, this strategy requires additional parameter tuning and input validation.

\paragraph*{Linkage pruning strategies}
Several alternative linkage hierarchy pruning strategies exist. For example, \citet{kpotufe2011pruning} propose pruning by density persistence, and \citet{stuetzle2010pruning} prune by excess of mass. HDBSCAN*'s EOM cluster selection strategy can also be viewed as a pruning strategy. However, it does not use a fixed EOM threshold. Instead, it removes branches if their parent has a larger relative excess of mass~\citep{campello2015hdbscan}. Consequently, it cannot be used as a pruning method for PLSCAN.

The pruning thresholds represent a cluster's size when pruning is viewed as a smoothing operation on the modelled density profile. Fundamentally, the choice between pruning by the number of observations or their density persistence determines whether one counts feature space without observations towards a cluster's size. We see a parallel to using $k$-nearest neighbour networks or $\e$-neighbour networks as manifold models in dimensionality reduction~\citep{belkin2003laplacianeigenmaps}. Future work could compare these alternative pruning strategies and formulate guidelines for their use.

\paragraph*{Usability study}
While we demonstrated PLSCAN's performance on several real-world datasets, a comparative usability study with clustering practitioners remains future work. Such a study is needed to compare the utility, usability, and generalisability of human--computer interfaces~\citep{crisan2018eval}. Specifically, it could determine whether analysing data with PLSCAN's cluster layers simplifies
pattern discovery compared to other clustering algorithms and data exploration workflows.

\subsection{Relations to prior work}%
\label{sec:discussion:related-work}

Several aspects of PLSCAN relate to prior work.

\paragraph*{Minimum cluster size filtration}
Perhaps most related to our work, \citet{neto2021multiscale} demonstrated how multiple HDBSCAN* cluster hierarchies can be computed for a range of minimum cluster size thresholds in a manner that is more efficient than computing each hierarchy independently. Notably, their approach sets $k = m_{\rm c}$, varying both parameters simultaneously. Consequently, the mutual reachability distances (Eq.~\ref{eq:mutual-reachability}) change in each hierarchy. In contrast, PLSCAN fixes $k$ and only varies $m_{\rm c}$. As a result, PLSCAN can efficiently detect and represent all topology-changing $m_{\rm c}$ thresholds in a single leaf-cluster hierarchy. That leaves $k$ as an input parameter. However, it has a predictable smoothing effect on the clustering results (Sec.~\ref{sec:demo:sensitivity}), making it easily tunable.

\paragraph*{Total persistence measures}
\citet{rieck2016clustering} previously used a total persistence measure to compare multiple clusterings based on the topological features they preserve. They defined a clustering's total persistence as the sum of squared cluster persistences. In contrast, PLSCAN simply sums the minimum cluster size persistences of clusters. Similar to optimising with Mean Squared Error (MSE) and Mean Absolute Error (MAE) losses, squaring persistence values emphasises clusterings with fewer but larger clusters. Our approach gives more weight to clusterings with more but smaller clusters.

Alternatively, \citet{leung2000scalespace} compute total persistence as a size-weighted mean. They evaluated individual cluster lifetimes as the logarithmic difference between their birth and death Gaussian kernel bandwidths, where the logarithm is used to model human perception~\citep{leung2000scalespace}. The conversions from distance to density in Eq.~\ref{eq:density-merge} and Eq.~\ref{eq:density} similarly compress large distance values. Future work could investigate whether the performance differences we observed between the distance and density persistence measures are related to this compressing effect.

\section*{Conclusion}%
\label{sec:conclusion}

This paper presents PLSCAN, a novel density-based clustering algorithm designed for exploratory data analysis. PLSCAN applies scale-space clustering principles to efficiently detect optimal minimum cluster size thresholds, thereby reducing the need for hyperparameter tuning. Consequently, it provides descriptive results that identify persistent high-density regions at multiple levels of detail, regardless of their shape and orientation. We compared PLSCAN to HDBSCAN* on several real-world datasets, demonstrating its higher average ARI and lower sensitivity to changes in the number of mutual reachability neighbours. In addition, we compared PLSCAN's stability and computational costs to those of HDBSCAN* and $k$-Means++, showing competitive stability and  run-times.

\section*{Acknowledgment}
Funding in direct support of this work: KU Leuven grant STG/23/040. We thank J.~F.~Bot for his review of early visualisations presenting our results.

\section*{Data Availability}
Data is available online at \url{https://doi.org/10.5281/zenodo.17974479}.

\bibliography{references.bib}
\end{multicols}

\appendix
\renewcommand{\thefigure}{\thesection\arabic{figure}}
\renewcommand{\thetable}{\thesection\arabic{table}}
\renewcommand{\thealgorithm}{\thesection\arabic{algorithm}}

\clearpage{}
\section{PLSCAN as persistent homology}%
\label{sec:appendix:tda}
\setcounter{table}{0}
\setcounter{figure}{0}
\setcounter{algorithm}{0}

\begin{figure*}[b!]
  \centering
  \hfill{}
  \begin{subfigure}{0.28\textwidth}
    \centering
    Condensed tree
    \includegraphics[width=\textwidth]{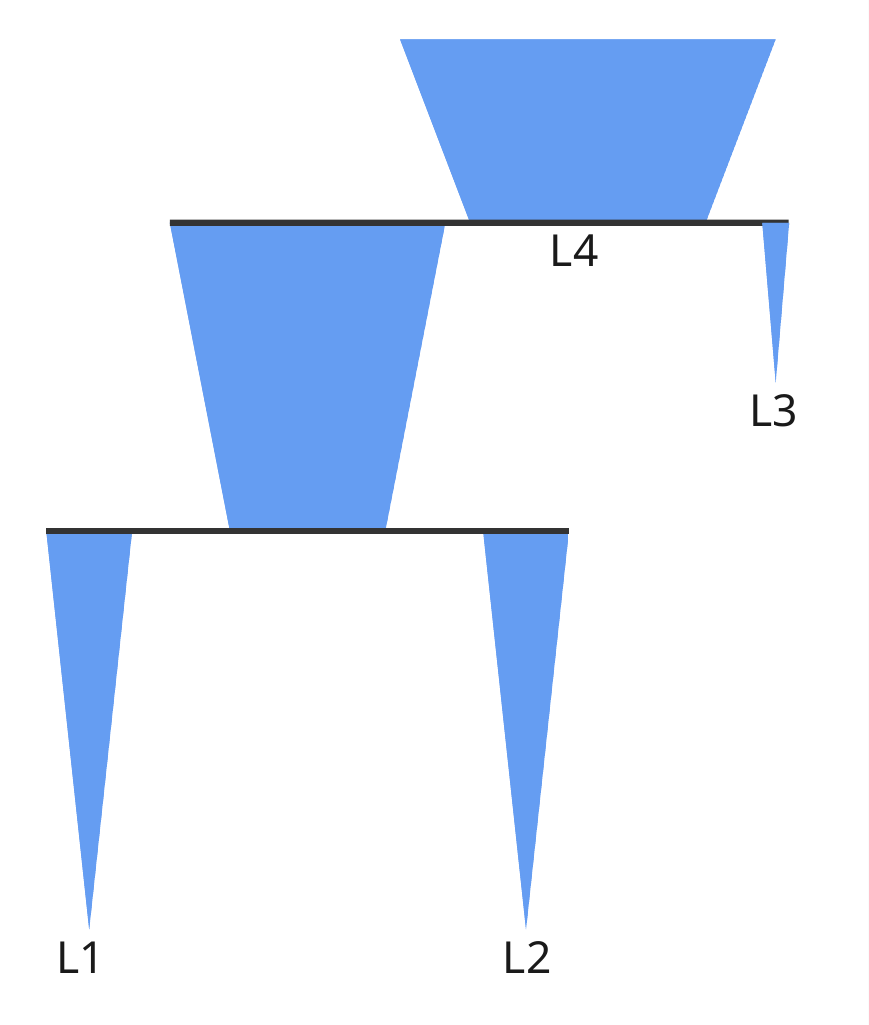}
    \caption{}%
    \label{fig:appendix:tda:early-deaths:condensed-tree}
  \end{subfigure}%
  \hfill{}
  \begin{subfigure}{0.28\textwidth}
    \centering
    Leaf tree
    \includegraphics[width=\textwidth]{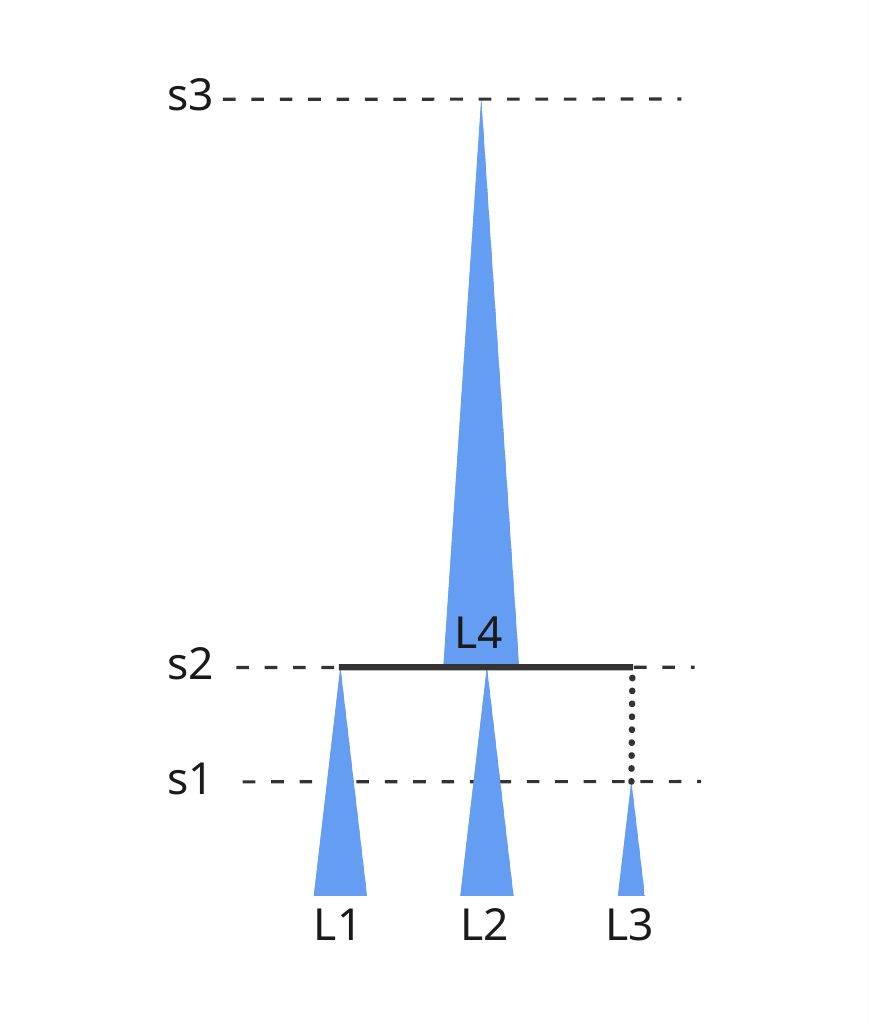}
    \caption{}%
    \label{fig:appendix:tda:early-deaths:leaf-tree}
  \end{subfigure}%
  \hfill{}
  \caption{Example of a leaf cluster ending before its parent appears in the filtration. (\textbf{a}) A schematic condensed tree with density along the $y$-axis and cluster sizes as icicle widths. (\textbf{b}) The corresponding schematic leaf-tree with minimum cluster sizes along the $y$-axis and excess of mass as icicle widths. Leaf \texttt{L3} contains fewer points than the descendants of its sibling \texttt{L1} and \texttt{L2}. Consequently, \texttt{L3} stops existing at $s_1$ before its parent \texttt{L4} appears at $s_2 > s_1$.}%
  \label{fig:appendix:tda:early-deaths}
\end{figure*}

\begin{multicols}{2}
\noindent This section defines a novel metric space for which zero-dimensional persistent homology uncovers the minimum cluster size range in which leaf-clusters exist. In other words, we describe a filtration that recovers the lifetimes in PLSCAN's leaf-cluster hierarchy.

\subsection{Preliminaries}%
\label{sec:appendix:tda:preliminaries}

Assume a linkage hierarchy has been constructed for a dataset $\vec{X} = \Set{\vec{x}_1, \dots, \vec{x}_{n}}$ with $n$ feature vectors $\vec{x}_i$ and an associated distance metric $\dist{\vec{x}_{i}, \vec{x}_{j}}$. Further, assume a monotonic parameterised pruning method filters out branches from the dendrogram. In other words, the higher the pruning parameter $s$, the more branches are removed. Let $\mathcal{L}_s = \Set{L_1, \dots, L_m}$ be the set of leaf clusters remaining after pruning the hierarchy with parameter $s$. HDBSCAN*'s minimum cluster size pruning meets these criteria and produces $\mathcal{L}_{m_{\rm c}} = \emptyset$ for $m_{\rm c} > n$. 

\subsection{Distance metric}%
\label{sec:appendix:tda:metric}

Next, we define a distance $\sdist{\vec{x}_i, \vec{x}_j}$ on $\vec{X}$ as the minimum pruning value $s$ that places two points in the same leaf cluster:
\begin{equation}
  \label{eq:pruning-metric}
  \sdist{\vec{x}_i, \vec{x}_j} = \setmin_s{\Set*{s \given \exists L \in \mathcal{L}_s : \vec{x}_i, \vec{x}_j \in L}},
\end{equation}
where $i \neq j$ and $\sdist{\vec{x}_i, \vec{x}_i} = 0$. The resulting distance is a metric on $\vec{X}$ and has a stronger-than-triangle-inequality: 
\begin{equation}
  \label{eq:ultrametric}
  \sdist{\vec{x}_i,\vec{x}_j} \leq \setmax{\Set*{\sdist{\vec{x}_i,\vec{x}_k}, \sdist{\vec{x}_k, \vec{x}_j}}}.
\end{equation}

Leaf clusters in $\mathcal{L}_s$ merge as the pruning parameter increases $s \to \infty$, forming a cluster hierarchy. Unlike in a linkage hierarchy, leaf clusters can disappear before their parent appears in the filtration (Fig.~\ref{fig:appendix:tda:early-deaths}). The current definition does not detect such disappearances, as it only measures when points are in the same leaf $L \in \mathcal{L}_s$. To detect these cases, we introduce special elements $\vec{\Pi} = \Set{\pi_i \given \forall \vec{x}_i \in \vec{X}}$ and update the metric for $\vec{Y} = \vec{X} \cup \vec{\Pi}$:
\begin{equation}
  \label{eq:pruning-noise}
  \sdist{\vec{x}_i, \pi_i} = \setmin_s{\Set*{s \given \forall L \in \mathcal{L}_s : \vec{x}_i \notin L}},
\end{equation}
with:
\begin{equation}
  \label{eq:pruning-noise-baseline}
  \sdist{\pi_i, \pi_i} = 0 \quad \text{and} \quad \sdist{\pi_i, \vec{x}_i} = \sdist{\vec{x}_i, \pi_i}.
\end{equation}
This definition constrains the distance at which points $\vec{x}_i$ connect to their $\pi_i$ to their respective leaf cluster $s$-lifetimes. Applying Eq.~\ref{eq:ultrametric}, the distance between elements in $\vec{\Pi}$ is:
\begin{equation}
    \label{eq:pruning-noise-dists}
    \sdist{\pi_i, \pi_j} = \setmax{\Set*{\sdist{\pi_i, \vec{x}_i}, \sdist{\vec{x}_i, \vec{x}_j}, \sdist{\vec{x}_j, \pi_j}}},
\end{equation}
where $i \neq j$.

\subsection{Persistent homology}%
\label{sec:appendix:tda:persistent-homology}

Applying persistent homology to the metric space $(\vec{Y}, \setsdist)$ uncovers the minimum cluster size ranges for which leaf-clusters exist. Connected components $CC$ in a Vietoris--Rips complex $VR_s$ with $\exists \vec{x}_i : \pi_i \notin CC$ form the set of leaves $\mathcal{L}_s$. The components merge either when points become noise (Eq.~\ref{eq:pruning-noise}) or when points become part of a single leaf (Eq.~\ref{eq:pruning-metric}). The special elements $\pi_i \in \vec{\Pi}$ do not introduce new connectivity between data points (Eq.~\ref{eq:pruning-noise-dists}). Consequently, they only contribute distances at which points become noise to the unique birth and death values in the barcode.

The leaf-tree construction in Sec.~\ref{sec:algorithm:leaf-tree} matches this formulation. The $s_{\rm max}$ values from Eq.~\ref{eq:leaf-tree:death-size} correspond to the distance at which the leaves merge (Eq.~\ref{eq:pruning-metric}) or one becomes noise (Eq.~\ref{eq:pruning-noise}). The $s_{\rm min}$ values from Eq.~\ref{eq:leaf-tree:birth-size} and parent assignments adapt the barcode to a nested hierarchical clustering, treating merges as new entities. In barcodes, the elder rule typically determines which component continues and which component disappears~\citep{edelsbrunner2010topology}.
\end{multicols}

\clearpage{}
\section{Per-dataset performance}%
\label{sec:appendix:performance}
\setcounter{table}{0}
\setcounter{figure}{0}
\setcounter{algorithm}{0}
\vfill{}
\begin{figure*}[!b]
  \centering
  \includegraphics[width=\textwidth]{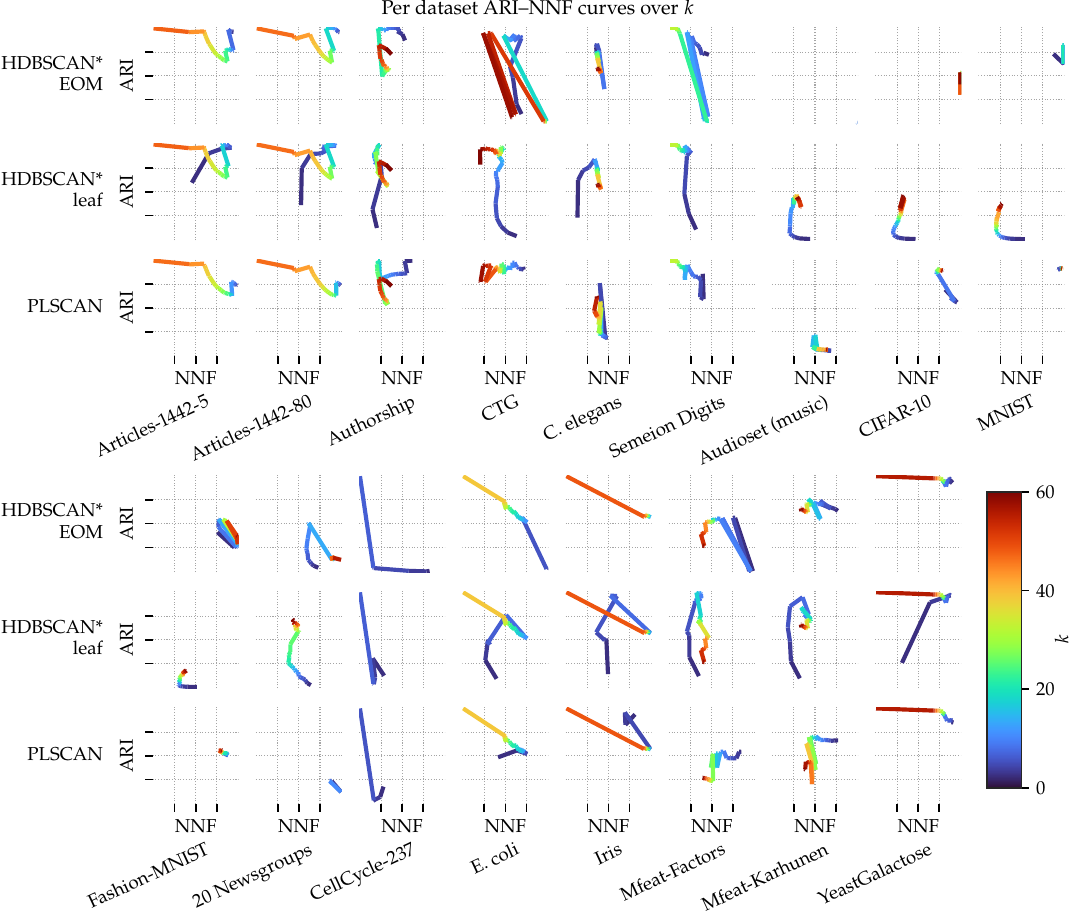}
  \caption{Per dataset ARI--NNF curves over $k$ for HDBSCAN* with EOM (top row), HDBSCAN* with leaf (middle row), and PLSCAN with size (bottom row). Colour encodes the value for $k$. Optimal performance corresponds to the top-right quadrants with high ARI and high NNF scores, indicating most points were part of a cluster and the clusters correspond to the ground-truth labels. The algorithms' sensitivity to $k$ is indicated by the compactness of the curves, with small variations corresponding to low sensitivity. Both $x$ and $y$ axes show values in $[0,1]$, with tick marks at $0.25$, $0.5$, and $0.75$.}%
  \label{fig:appendix:performance}
\end{figure*}
\vfill{}

\begin{multicols}{2}
  \noindent{}This appendix shows per-dataset performances for the sensitivity benchmark in Sec.~\ref{sec:demo:sensitivity}. Figure~\ref{fig:appendix:performance} visualises the ARI--NNF curves over $k$ for each dataset and algorithm. The plots provide information about which datasets each algorithm performed well on and how the performance changed with respect to $k$. Notice, for example, that HDBSCAN* with the leaf cluster selection strategy (middle row) has larger blue traces at lower ARI and NNF values, indicating it had worse scores at low $k$ values. Additionally, the traces' spreads indicate how much performance varied over $k$ on each dataset. Compare, for example, the traces for PLSCAN and HDBSCAN* EOM on the Fashion-MNIST and CIFAR-10 datasets.
\end{multicols}
\vfill{}

\clearpage{}
\section{Pseudocode}%
\label{sec:appendix:plscan}
\setcounter{table}{0}
\setcounter{figure}{0}
\setcounter{algorithm}{0}
\begin{algorithm*}[!htb]
  \caption{Our algorithm for computing the condensed cluster tree. Where \citet{mcinnes2017accelerated} perform a breadth-first traversal to collect all points in pruned branches, we track where those points should be written. This allows us to perform a single loop over the linkage tree, writing each data point as we encounter it. This is possible because our linkage tree tracks weighted and unweighted sizes in \texttt{child\_size} and \texttt{child\_count}, respectively.}%
  \label{alg:condense-tree}
  \begin{algorithmic}[1]
    \Function{condense\_tree}{\texttt{linkage\_tree}, \texttt{spanning\_tree}, \texttt{sample\_weights}, $n$, $m_c$} 
      \LComment{Simplifies the \texttt{linkage\_tree} and
      \texttt{spanning\_tree} keeping only clusters with a (weighted) minimum
      cluster size $m_{\rm c}$, where $n$ is the total number of data points.} 
      \State \texttt{condensed\_tree} $\gets$ initialize with empty \texttt{parent\_id}, \texttt{child\_id}, \texttt{distance}, \texttt{child\_size} arrays
      \State \texttt{index} $\gets$ 0                                                                              \Comment{First available row in the \texttt{condensed\_tree}.}
      \State \texttt{next\_label} $\gets$ $n$                                                                      \Comment{Next available cluster label.}
      \State \texttt{parent\_of} $\gets$ initialize $n-1$ array with value $n$                                     \Comment{Used to track condensed tree parent labels.}
      \State \texttt{pending\_idx} $\gets$ initialize $n-1$ array with empty values                                \Comment{Used to reserve rows in the condensed tree.} 
      \State \texttt{pending\_distance} $\gets$ initialize $n-1$ array with empty values                           \Comment{Used to propagate ancestor distances.}
      \For{each \texttt{link\_idx} in \texttt{linkage\_tree}}                                                      \Comment{In descending distance order.}
        \State \texttt{out\_idx}, \texttt{distance} $\gets$ \Call{update\_output\_index}{\texttt{link\_id}, $m_c$}
        \State \texttt{out\_idx} $\gets$ \Call{write\_or\_delay}{\texttt{linkage\_tree.parent\_id[link\_idx]}, \texttt{distance}, \texttt{out\_idx}, $n$, $m_c$}
        \State \texttt{out\_idx} $\gets$ \Call{write\_or\_delay}{\texttt{linkage\_tree.child\_id[link\_idx]}, \texttt{distance}, \texttt{out\_idx}, $n$, $m_c$}
        \State \texttt{index}, \texttt{next\_label} $\gets$ \Call{write\_merge}{\texttt{link\_idx}, \texttt{index}, \texttt{next\_label}, $n$, $m_c$}
      \EndFor{}
      \Return \texttt{condensed\_tree}
      \EndFunction
      \\
      \Function{update\_output\_index}{\texttt{link\_idx}, $m_c$}
      \LComment{Decides where data-point rows are written. Unlisted
      variables are mutably captured from \Call{condense\_tree}{}.}
        \If{\texttt{linkage\_tree.child\_size[link\_idx]} $<$ $m_c$}                            
          \State \texttt{out\_idx} $\gets$ \texttt{parent\_of}[\texttt{link\_idx}]                      \Comment{Write pruned branches to reserved spots.}
          \State \texttt{distance} $\gets$ \texttt{pending\_distance}[\texttt{link\_idx}]               \Comment{Use ancestor distance for pruned rows.}
        \Else
          \State \texttt{out\_idx} $\gets$ \texttt{index}                                               \Comment{Write accepted branches to next available row.}
          \State \texttt{distance} $\gets$ \texttt{spanning\_tree.distance[link\_idx]}                  \Comment{Use current distance for accepted rows.}
          \State \texttt{index} $\gets$ \texttt{index} + number of pruned points in \texttt{link\_idx}  \Comment{Reserve rows for potential pruned descendants.}
        \EndIf
        \Return \texttt{out\_idx}, \texttt{distance}
      \EndFunction
      \\
      \Function{write\_or\_delay}{\texttt{link\_id}, \texttt{distance}, \texttt{out\_idx}, $n$, $m_c$}
      \LComment{Writes or propagates rows in the linkage tree. Unlisted
      variables are mutably captured from \Call{condense\_tree}{}.}
      \If{\texttt{link\_id} $<$ $n$}                                                                                   \Comment{Write a row for single data points.}
         \State \texttt{condensed\_tree[out\_idx++]} $\gets$ \{\texttt{parent\_of[link\_idx]}, \texttt{link\_id}, \texttt{distance}, \texttt{sample\_weights[link\_id]}\}
      \Else
         \State \texttt{parent\_of[link\_id $-$ $n$]} $\gets$ \texttt{parent\_of[link\_idx]}                           \Comment{Propagate new parent label.}
         \If{\texttt{linkage\_tree.child\_size[link\_id $-$ $n$]} $<$ $m_c$}
            \State \texttt{pending\_idx[link\_id $-$ $n$]} $\gets$ \texttt{out\_idx}                                   \Comment{Propagate reserved row index.}
            \State \texttt{pending\_distance[link\_id $-$ $n$]} $\gets$ \texttt{distance}                              \Comment{Propagate ancestor distance.}
            \State \texttt{out\_idx} $\gets$ \texttt{out\_idx} + \texttt{linkage\_tree.child\_count[link\_id $-$ $n$]} \Comment{Reserve rows for all descendants.}
         \EndIf
      \EndIf
      \Return \texttt{out\_idx}
      \EndFunction
      \\
      \Function{write\_merge}{\texttt{link\_idx}, \texttt{index}, \texttt{next\_label}, $n$, $m_c$}
      \LComment{Writes rows for cluster merges. Unlisted variables are mutably captured from \Call{condense\_tree}{}.}
      \If{parent and child sides of \texttt{link\_idx} contain at least $m_c$ points}
         \State \texttt{parent} $\gets$ \texttt{parent\_of[link\_idx]}
         \If{\texttt{parent} = $n$}
            \State \texttt{parent} $\gets$ ++\texttt{next\_label} \Comment{Adjust for phantom root.}
         \EndIf
         \State \texttt{parent\_idx} $\gets$ \texttt{linkage\_tree.parent\_id[link\_idx] $-$ $n$}    \Comment{Safe to subtract $n$ here because $m_c$ must be}
         \State \texttt{child\_idx} $\gets$ \texttt{linkage\_tree.child\_id[link\_idx] $-$ $n$}      \Comment{greater than the maximum sample weight.}
         \State \texttt{parent\_of[parent\_idx]} $\gets$ ++\texttt{next\_label}                      \Comment{Introduce new parent label.}
         \State \texttt{condensed\_tree[index++]} $\gets$ \{\texttt{parent}, \texttt{next\_label}, \texttt{distance}, \texttt{linkage\_tree.child\_size[parent\_idx]}\}
         \State \texttt{parent\_of[child\_idx]} $\gets$ ++\texttt{next\_label}                       \Comment{Introduce new parent label.}
         \State \texttt{condensed\_tree[index++]} $\gets$ \{\texttt{parent}, \texttt{next\_label}, \texttt{distance}, \texttt{linkage\_tree.child\_size[child\_idx]}\}
      \EndIf
      \Return \texttt{index}, \texttt{next\_label}
      \EndFunction
  \end{algorithmic}
\end{algorithm*}

\begin{algorithm*}[!htb]
  \caption{This leaf tree algorithm converts a condensed tree to the leaf tree format. Leaf trees lists properties of all parent-IDs in the condensed tree (i.e., segments in Fig.~\ref{fig:algorithm:leaf-tree:condensed-trees}). Specifically, we track the parent and min/max size and distance for which segments are a leaf-clusters. The parent IDs are re-normalised to start at $0$ and index into the leaf tree, facilitating traversal. Segments that do not become leaf can be recognized by their minimum sizes being larger than their maximum sizes. The other segments form a minimum cluster size barcode. Our condensed tree stores which rows corresponding to the cluster hierarchy, allowing for fast access to sibling segments.}%
  \label{alg:leaf-tree}
  \begin{algorithmic}[1]
    \Function{leaf\_tree}{\texttt{condensed\_tree}, $n$, $k$}
      \LComment{Extracts leaves from the \texttt{condensed\_tree} where $n$ is
      the total number of data points and $k$ the number of neighbours used for
      the mutual reachability distance.}
      \State \texttt{leaf\_tree} $\gets$ allocate for \texttt{condensed\_tree.child\_id.max() $-$ $n$} segments          \Comment{Max is at last cluster row.}
      \State \texttt{leaf\_tree.parent[:]} $\gets$ $0$                                                                   \Comment{Initialize with phantom root $0$.}
      \State \texttt{leaf\_tree.max\_size[0]} $\gets$ $n$                                                                \Comment{Initialize phantom root with largest size.}
      \State \texttt{leaf\_tree.min\_size[:]} $\gets$ $k$                                                                \Comment{Initialize with smallest size.}
      \State \texttt{leaf\_tree.max\_dist[:]} $\gets$ \texttt{condensed\_tree.distance[0]}                               \Comment{Initialize with highest distance.}
      \State \texttt{leaf\_tree.min\_dist[condensed\_tree.parent\_id $-$ $n$]} $\gets$ \texttt{condensed\_tree.distance} \Comment{Last value per parent.}
      \For{each cluster row pair \texttt{left\_idx}, \texttt{right\_idx} in the \texttt{condensed\_tree}}                \Comment{Increasing distance order.}
        \State \texttt{out\_idx} $\gets$ \texttt{condensed\_tree.child\_id[left\_idx] $-$ $n$}                           \Comment{Get child leaf tree index.}
        \State \texttt{parent\_idx} $\gets$ \texttt{condensed\_tree.parent\_id[left\_idx] $-$ $n$}                       \Comment{Get parent leaf tree index.}
        \State \texttt{size} $\gets$ \texttt{condensed\_tree.child\_size[left\_idx, right\_idx].min()}                   \Comment{Both siblings die when one dies.}
        \State \texttt{leaf\_tree.parent\_id[out\_idx, out\_idx $-$ $1$]} $\gets$ \texttt{parent\_idx}                   \Comment{Sibling are always adjacent.}
        \State \texttt{leaf\_tree.max\_dist[out\_idx, out\_idx $-$ $1$]} $\gets$ \texttt{condensed\_tree.distance[left\_idx]} 
        \State \texttt{leaf\_tree.max\_size[out\_idx, out\_idx $-$ $1$]} $\gets$ \texttt{size}                              
        \State \texttt{leaf\_tree.min\_size[parent\_idx]} $\gets$ ${\rm max}($                                           \Comment{Propagate minimum size upwards.}
        \State \hspace{\algorithmicindent} \texttt{size}, \texttt{leaf\_tree.min\_size[out\_idx]}, \texttt{leaf\_tree.min\_size[out\_idx $-$ $1$]} 
        \State $)$ 
        \If{\texttt{leaf\_tree.parent[parent\_idx]} is $0$}                                                              \Comment{Also propagate to phatom root.}   
          \State \texttt{leaf\_tree.min\_size[0]} $\gets$ ${\rm max}($\texttt{leaf\_tree.min\_size[0]}, \texttt{leaf\_tree.min\_size[parent\_idx]}$)$
        \EndIf
      \EndFor
      \State \texttt{leaf\_tree.max\_size[leaf\_tree.parent == 0]} $\gets$ \texttt{leaf\_tree.min\_size[0]}              \Comment{Largest observed.}
      \State \Return \texttt{leaf\_tree}
    \EndFunction
  \end{algorithmic}
\end{algorithm*}
\begin{algorithm*}[!htb]
  \caption{The persistence trace aggregates leaf cluster persistences over their minimum cluster size lifetimes. Alternatively, bi-persistence measures that also consider the clusters' distance or density persistence can be used in place of the size persistence. These measures require listing all points that are part of the leaf cluster throughout its size lifetime. Leaf clusters can contain points not in their condensed tree segment because siblings can vary in size! The correct points and their distances can be found by passing points up the leaf tree and tracking the clusters' sizes. This is also why the min.~and max.~distances in the leaf tree do not match their leaf cluster's distance-lifetime. Instead, they simply describe the condensed-tree segment's distance persistence!}%
  \label{alg:persistence-trace}
  \begin{algorithmic}[1]
    \Function{persistence\_trace}{\texttt{leaf\_tree}, \texttt{condensed\_tree}, $n$}
      \LComment{Computes the total mininum cluster size persistences trace over all
      leaf clusters that exist at all size thresholds.}
      \State \texttt{min\_size} $\gets$ unique values in \texttt{leaf\_tree.min\_size} and \texttt{leaf\_tree.max\_size} \Comment{Supports weighted samples.}
      \State \texttt{persistence} $\gets$ zero array matching size of \texttt{min\_size}
      \For{each \texttt{idx} in \texttt{leaf\_tree}}
        \State \texttt{birth} $\gets$ \texttt{leaf\_tree.min\_size[idx]}
        \State \texttt{death} $\gets$ \texttt{leaf\_tree.max\_size[idx]}
        \State \texttt{persistence[birth $\le$ min\_size $<$ death]} $\gets$ add \texttt{death} $-$ \texttt{birth}       \Comment{Binary search for the indices.}
      \EndFor
      \State \Return \texttt{min\_size}, \texttt{persistence}
    \EndFunction
  \end{algorithmic}
\end{algorithm*}
\begin{algorithm*}[!htb]
  \caption{Our cluster selection algorithm computes a weighted flat clustering using a straight minimum cluster size cut in the leaf tree. Like EOM, this produces clusters with varying densities. Rather than re-running a union--find algorithm like \citet{mcinnes2017accelerated}, we efficiently assign cluster labels to each leaf tree segment. Afterwards we can read cluster labels from condensed tree parent IDs in a single iteration.}%
  \label{alg:select-clusters}
  \begin{algorithmic}[1]
    \Function{select\_clusters}{\texttt{trace}, \texttt{leaf\_tree}, \texttt{condensed\_tree}, $n$}
      \LComment{Extracts cluster labels and membership probabilities for the
      minimum cluster size that maximizes the total persistence in
      \texttt{trace}. The \texttt{condensed\_tree} stores which parents data
      points belong to. The \texttt{leaf\_tree} store which parents exist at the
      optimal size threshold.}
      \State \texttt{cut\_size} $\gets$ \texttt{trace.min\_size[$\argmax{\texttt{trace.persistence}}$]}
      \State \texttt{selected} $\gets$ indices where \texttt{leaf\_tree.min\_size} $\le$ \texttt{cut\_size} $\le$ \texttt{leaf\_tree.max\_size}
      \State \texttt{segment\_labels} $\gets$ \Call{compute\_segment\_labels}{\texttt{leaf\_tree}, \texttt{selected}}
      \State \Return \Call{apply\_segment\_labels}{\texttt{leaf\_tree}, \texttt{condensed\_tree}, \texttt{selected}, \texttt{segment\_labels}, $n$}
    \EndFunction
    \\
    \Function{compute\_segment\_labels}{\texttt{leaf\_tree}, \texttt{selected}}
      \LComment{Lists the cluster label for each leaf tree segment. This
      approach works because the condensed tree, and thereby the leaf tree, are
      ordered by distance! So all leaf tree segements before a selected segments
      are above it in the tree.}
      \State \texttt{segment\_labels} $\gets$ allocate for \texttt{leaf\_tree.size()} segments
      \State \texttt{segment\_labels[0]} $\gets$ $-1$                                                           \Comment{Phantom root is always noise.}
      \State \texttt{label} $\gets$ $0$
      \For{each \texttt{idx} in \texttt{segment\_labels}}                                                       \Comment{Starting at \texttt{idx}$=1$.}
        \If{\texttt{label} $<$ \texttt{selected.size()} \textbf{and} \texttt{selected[label]} $=$ \texttt{idx}} \Comment{Found the next selected segment.}
          \State \texttt{segment\_labels[idx]} $\gets$ \texttt{label++}                                         \Comment{Bump the label.}
        \Else
          \State \texttt{segment\_labels[idx]} $\gets$ \texttt{segment\_labels[leaf\_tree.parent\_id[idx]]}     \Comment{Inherit label from parent.}
        \EndIf
      \EndFor
      \State \Return \texttt{segment\_labels}
    \EndFunction
    \\
    \Function{apply\_segment\_labels}{\texttt{leaf\_tree}, \texttt{condensed\_tree}, \texttt{selected}, \texttt{segment\_labels}, $n$}
      \LComment{Lists data point cluster labels and membership probabilities.}
      \State \texttt{labels} $\gets$ array of size $n$ initialized to $-1$
      \State \texttt{probabilities} $\gets$ array of size $n$ initialized to $0.0$
      \For{each row \texttt{idx} in the \texttt{condensed\_tree}}
        \State \texttt{child\_id} $\gets$ \texttt{condensed\_tree.child\_id[idx]}
        \If{\texttt{child\_id} $\ge$ $n$}                                                                        \Comment{Skip cluster rows.}
          \State \textbf{continue}
        \EndIf
        \State \texttt{label} $\gets$ \texttt{segment\_labels[\texttt{condensed\_tree.parent\_id[idx]} $-$ $n$]} \Comment{Get label from parent segment.}                 
        \State \texttt{labels[child\_id]} $\gets$ \texttt{label}
        \If{\texttt{label} $\ge 0$}
          \State \texttt{probabilities[child\_id]} $\gets$ ${\rm min}(1$,                                        \Comment{Clamp probability to $[0, 1]$.}
          \State \hspace{\algorithmicindent} $($\texttt{leaf\_tree.max\_dist[selected[label]]} $-$ \texttt{condensed\_tree.distance[idx]} $)$ $/$
          \State \hspace{\algorithmicindent} $($\texttt{leaf\_tree.max\_dist[selected[label]]} $-$ \texttt{leaf\_tree.min\_dist[selected[label]]} $)$
          \State $)$
        \EndIf
      \EndFor
      \State \Return \texttt{labels}, \texttt{probabilities}
    \EndFunction
  \end{algorithmic}
\end{algorithm*}
\clearpage{}

\begin{multicols}{2}
\end{document}